\documentclass[letterpaper]{article} % DO NOT CHANGE THIS
\usepackage{aaai23}  % DO NOT CHANGE THIS
\usepackage{times}  % DO NOT CHANGE THIS
\usepackage{helvet}  % DO NOT CHANGE THIS
\usepackage{courier}  % DO NOT CHANGE THIS
\usepackage[hyphens]{url}  % DO NOT CHANGE THIS
\usepackage{graphicx} % DO NOT CHANGE THIS
\usepackage{natbib}  % DO NOT CHANGE THIS AND DO NOT ADD ANY OPTIONS TO IT
\usepackage{caption} % DO NOT CHANGE THIS AND DO NOT ADD ANY OPTIONS TO IT
\usepackage{algorithm}
\usepackage{algorithmic}

\usepackage{tabularx}

\usepackage{amsmath}
\usepackage{booktabs}
\usepackage{subfigure}
\usepackage{bm}
\usepackage{amssymb}
\usepackage{multirow}
\usepackage{comment}
\newcolumntype{C}[1]{>{\centering\let\newline\\\arraybackslash\hspace{0pt}}m{#1}}

\usepackage{newfloat}
\usepackage{listings}
\DeclareCaptionStyle{ruled}{labelfont=normalfont,labelsep=colon,strut=off} % DO NOT CHANGE THIS
\floatstyle{ruled}
\newfloat{listing}{tb}{lst}{}
\floatname{listing}{Listing}
\title{Imperceptible Adversarial Attack via Invertible Neural Networks}
\author{
    Zihan Chen,
    Ziyue Wang,
    Junjie Huang\thanks{Corresponding author},
    Wentao Zhao,
    Xiao Liu,
    Dejian Guan
}
\affiliations{
    College of Computer Science, National University of Defense Technology, Changsha, Hunan, China\\
    \{chenzihan21, wangzy13, jjhuang, wtzhao, liuxiao13a, guandejian20\}@nudt.edu.cn 
}

\usepackage{bibentry}
\begin{document}

\maketitle

\begin{abstract}
Adding perturbations via utilizing auxiliary gradient information or discarding existing details of the benign images are two common approaches for generating adversarial examples.
Though visual imperceptibility is the desired property of adversarial examples, conventional adversarial attacks still generate traceable adversarial perturbations. 
In this paper, we introduce a novel {Adv}ersarial Attack via {I}nvertible {N}eural {N}etworks (AdvINN) method to produce robust and imperceptible adversarial examples. Specifically, AdvINN fully takes advantage of the information preservation property of Invertible Neural Networks and thereby generates adversarial examples by simultaneously adding class-specific semantic information of the target class and dropping discriminant information of the original class. Extensive experiments on CIFAR-10, CIFAR-100, and ImageNet-1K demonstrate that the proposed AdvINN method can produce less imperceptible adversarial images than the state-of-the-art methods and AdvINN yields more robust adversarial examples with high confidence compared to other adversarial attacks. Code is available at \url{https://github.com/jjhuangcs/AdvINN}.

% Code is available at. 

\end{abstract}

\section{Introduction}

%-----------------------------------------------------------
\begin{figure}[!h]
    \centering
    \includegraphics[width=0.85\linewidth]{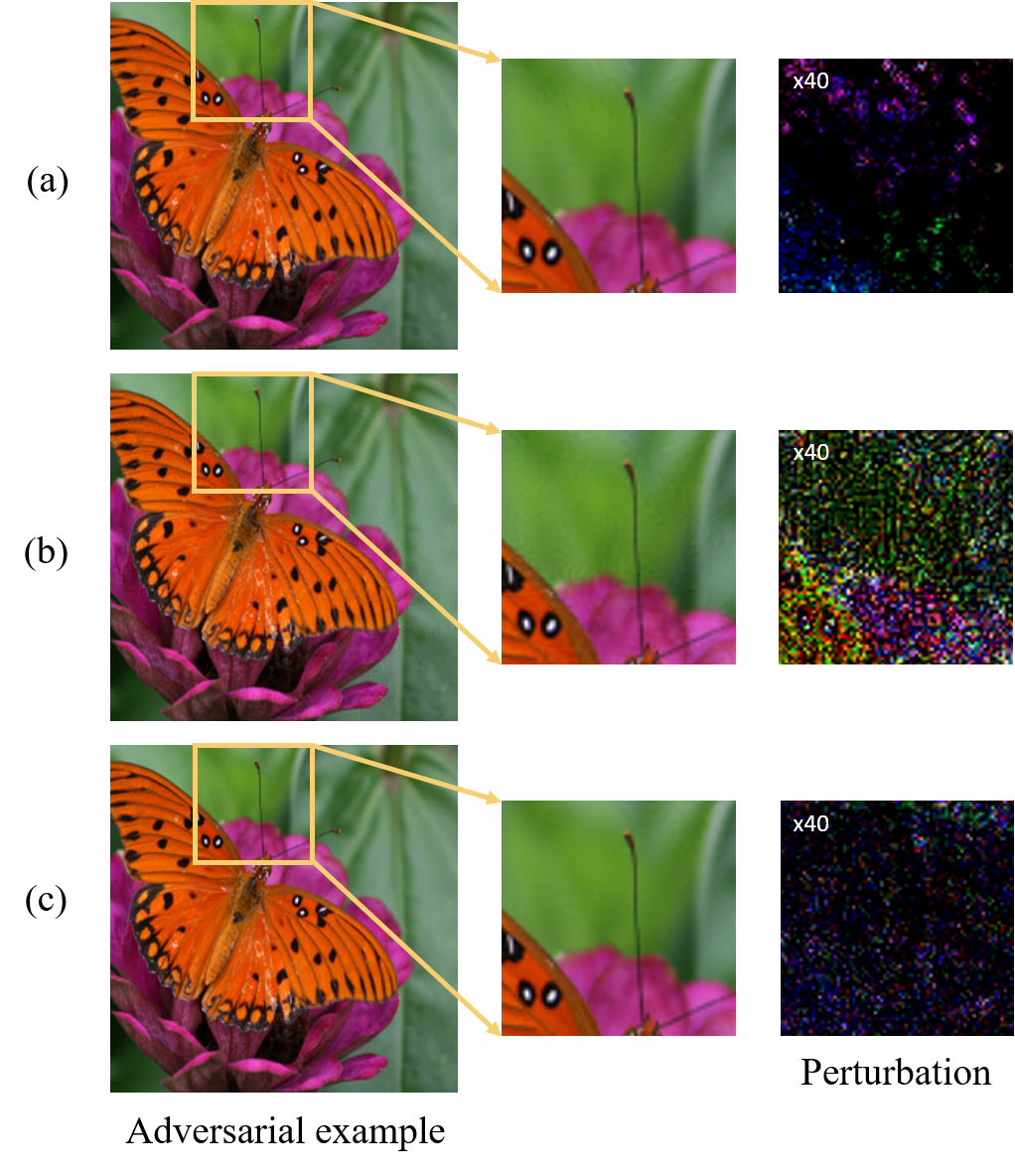}
    % \subfigure[]{
    % \begin{minipage}[b]{0.4\textwidth}
    % \includegraphics[width=0.45\linewidth]{image/Flower_PerCAL_img.png}\vspace{0.01pt}
    % \includegraphics[width=0.45\linewidth]{image/Flower_PerCAL_adv.png}
    % \end{minipage}
    % }

    % \subfigure[]{
    % \begin{minipage}[b]{0.4\textwidth}
    % \includegraphics[width=0.45\linewidth]{image/Flower_SSAH_img.png}\vspace{0.01pt}
    % \includegraphics[width=0.45\linewidth]{image/Flower_SSAH_adv.png}
    % \end{minipage}
    % }

    % \subfigure[]{
    % \begin{minipage}[b]{0.4\textwidth}
    % \includegraphics[width=0.45\linewidth]{image/Flower_AdvINN_img.png}\vspace{0.01pt}
    % \includegraphics[width=0.45\linewidth]{image/Flower_AdvINN_adv.png}
    % \end{minipage}
    % }
    \caption{Adversarial examples generated by (a) PerC-AL~\cite{zhao2020towards}, (b) SSAH~\cite{luo2022frequency}, and (c) the proposed AdvINN, respectively. All perturbations are enlarged for better visualization.}
    \label{fig:first_look}
\end{figure}
%-----------------------------------------------------------
Deep Neural Networks (DNNs) have achieved outstanding performance in a wide range of applications, however, have shown to be vulnerable to adversarial examples~\cite{DBLP:journals/corr/SzegedyZSBEGF13,goodfellow2014explaining,Akhtar2018ThreatOA,Hendrycks2021NaturalAE}. By adding mild adversarial noise to a benign image, classification DNNs can be easily deceived and misclassify this adversarial example to an erroneous class label. Though the existence of adversarial examples may hinder the applications of DNNs to risk sensitive domains, it further promotes investigation on robustness of DNNs.

Adversarial examples can be generated by either adding or dropping certain information with respect to the input benign images. Adding adversarial perturbations~\cite{DBLP:journals/corr/SzegedyZSBEGF13,moosavi2016deepfool,carlini2017towards} to clean images is the most common approach to craft adversarial examples. Fast Gradient Sign Method (FGSM)~\cite{DBLP:journals/corr/SzegedyZSBEGF13} and its variations~\cite{kurakin2016adversarial,dong2018boosting,lin2019nesterov} add adversarial noise to the benign image according to the sign of the gradients of the loss function with respect to the input image. An alternative is to mix a sequence of images to make the classifier output erroneous predictions and improve the transferability of generated adversarial examples~\cite{wang2021admix}, since information from images of other classes could disturb the prediction of DNNs.
Recently, dropping existing information from the original images has also shown to be an effective way to generate adversaries~\cite{duan2021advdrop}. 
Compared to methods of adding adversarial perturbations to the benign images, AdvDrop~\cite{duan2021advdrop} shows stronger robustness against denoising-based defence methods and will not lead to suspicious increase of image storage size.

%-----------------------------------------------------------
\begin{figure*}[t]
    \centering
    \includegraphics[width=0.78\linewidth]{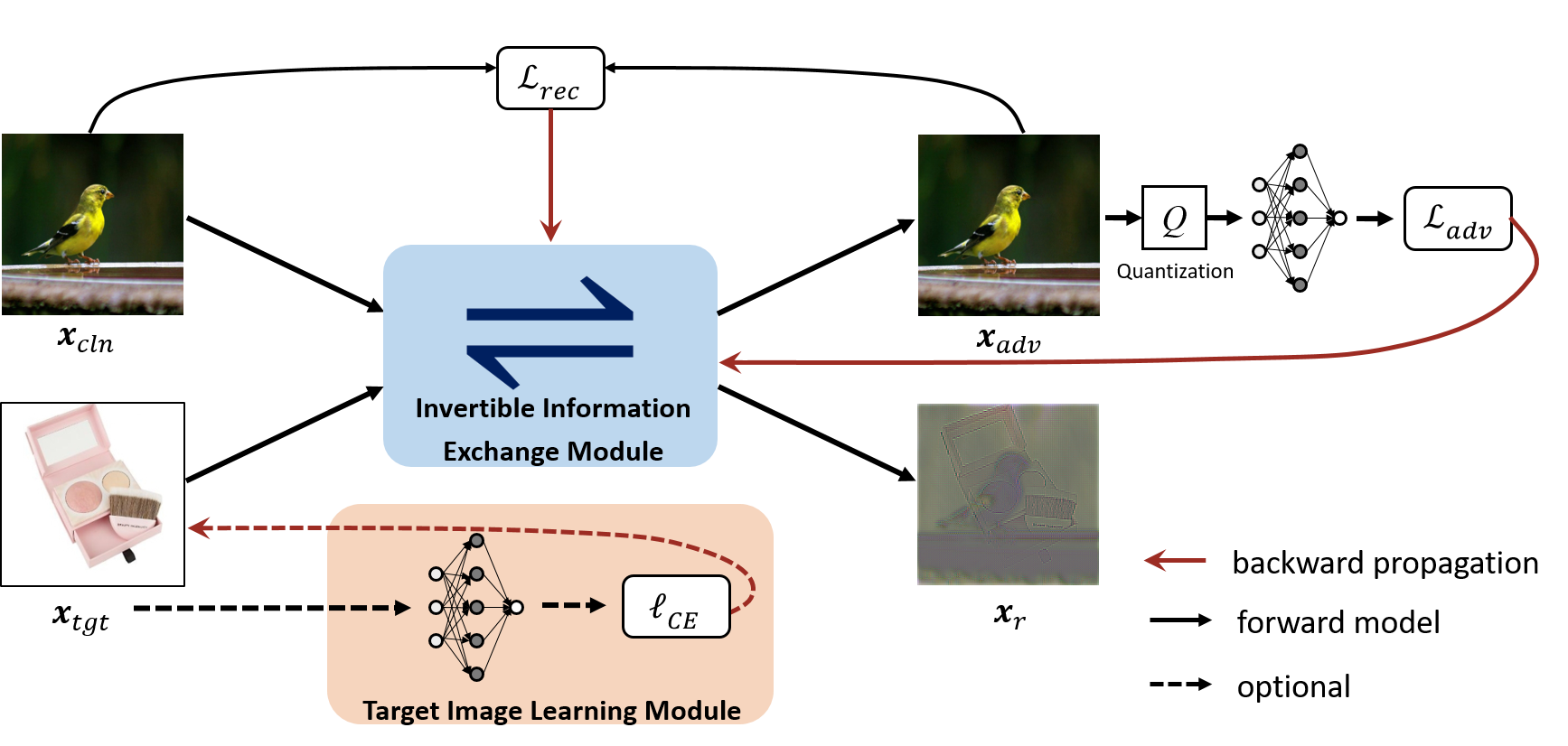}
    \caption{The overview architecture of our proposed Adversarial Attack using Invertible Neural Networks (AdvINN) method. The Invertible Information Exchange Module, which is with the information preservation property, non-linearly exchanges information between the input benign image and the target image. The Target Image Learning Module is used to update the learnable target image $x_{tgt}$. The quantization module is set to round the pixel values of the generated adversarial examples $x_{adv}$ to be integers and within the range of [0,255].}
    \label{fig:overview}
\end{figure*}
%-----------------------------------------------------------

Adversarial examples crafted by adding or dropping information are both able to deceive DNNs with incorrect prediction of image contents, however, both approaches have their limitations. The methods based on adding adversarial perturbations may lead to perceptible noise patterns and noticeable increase of image storage size, while the method of dropping existing information has limited performance on targeted attacks. Therefore, it is of great potential to make an attempt to combine the best features from two perspectives by simultaneously adding semantic information from the target image and dropping semantic information of the original class to craft adversarial examples.

In this paper, we propose a novel Adversarial attack method using Invertible Neural Networks, termed AdvINN, by leveraging the information preservation property of Invertible Neural Networks (INNs) to achieve simultaneously adding extra information and dropping existing details. 
Specifically, given a clean image, a target image is selected or learned as the source of information for adding adversarial perturbations. The clean image and the target image are inputs to an Invertible Information Exchange Module (IIEM) for alternating update. The amount of information within the input and output of IIEM keeps the same due to its information preservation property.
Therefore, driven by an adversarial loss and a reconstruction loss, the generated adversarial image will gradually transfer discriminant features of the clean image to the residual image and at the same time add class-specific semantic features from the target image to form an adversarial example.

The contribution of this paper is three-fold:
\begin{itemize}
    \item We propose a novel Adversarial attack method using Invertible Neural Networks {(AdvINN)} which exploits the information preservation property of Invertible Neural Networks and is able to achieve simultaneously adding class-specific information from a target image and dropping semantic information of the original class.

    \item We propose three approaches to choose the target image, including highest confidence image, universal adversarial perturbation, and learnable classifier guided target image. With the proposed AdvINN, class-specific features can be effectively transferred to the input image leading to highly interpretable and imperceptible results.
    
    \item With comprehensive experiments and analysis, we have demonstrated the effectiveness and robustness of the proposed AdvINN method, and shown that the adversarial examples generated by AdvINN are more imperceptible and with high attacking success rates.
\end{itemize}

\section{Related Works}
\label{sec:related work}
\subsection{Adversarial Attack}

Adversarial attacks~\cite{DBLP:journals/corr/SzegedyZSBEGF13} aim to deceive DNNs with adversarial examples whose difference to the input benign image is bounded by $l_{\infty}$-norm. 
That is, an adversarial example should be able to fool DNNs and at the same time be as imperceptible as possible. 
In general, adversarial examples can be crafted by adding disturbing adversarial perturbations to clean images or dropping crucial information from the clean images.

\textbf{Adding adversarial perturbations} to clean images is a predominant way to generate adversarial examples in recent works.
FSGM~\cite{DBLP:journals/corr/SzegedyZSBEGF13} proposes to add adversarial perturbation in the direction of sign of gradient. 
BIM~\cite{kurakin2016adversarial} increases the number of iterations and updates with smaller steps to improve the attacking success rate. 
StepLL~\cite{kurakin2016adversarial} proposes to choose the least-likely class as the target class and can generate adversarial examples which are highly misclassified by Inceptionv3~\cite{szegedy2016rethinking}. 
PGD~\cite{madry2018towards} is similar to BIM, while with the randomly initialized starting point in the neighborhood of ground-truth image. 
DeepFool~\cite{moosavi2016deepfool} proposes to add the minimal norm adversarial perturbation around decision boundary to make false predictions.
C\&W~\cite{carlini2017towards} attempts to find a balance between imperceptible perturbations and adversarial attacks with $l_{0}$, $l_{2}$ and $l_{\infty}$-norm regularizations. 
\cite{wang2021admix} propose {Admix} method to generate more transferable adversarial examples by mixing the input image with a small portion of images sampled from other categories.

%-----------------------------------------------------------
\begin{figure}[t]
    \centering
    \includegraphics[height=0.35\linewidth]{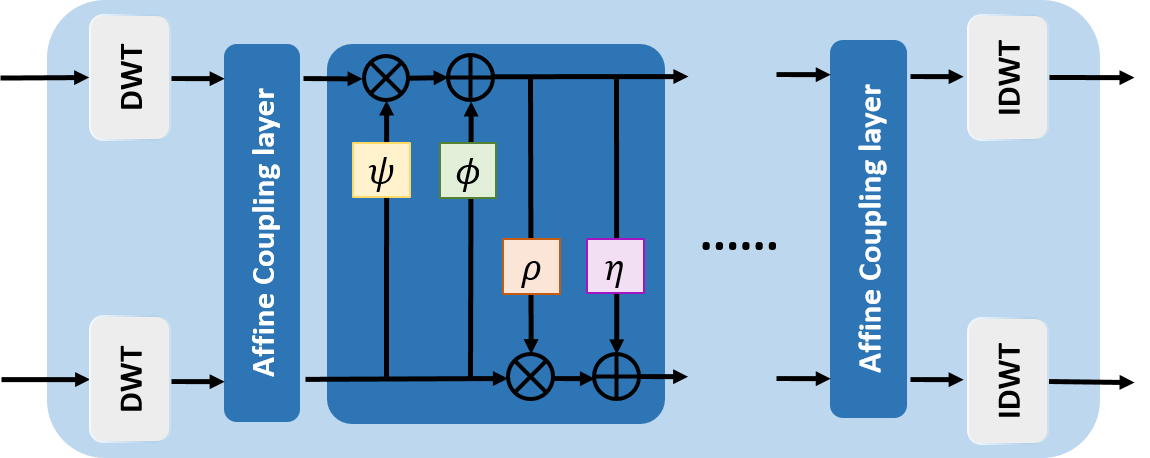}
    \caption{The network architecture of the Invertible Information Exchange Module. }
    \label{fig:IIEM}
\end{figure}
%-----------------------------------------------------------

\textbf{Dropping existing information}
has also proven to be able to successfully craft adversarial examples, which provides a new perspective for generating adversarial examples. \cite{duan2021advdrop} propose AdvDrop method which learns the quantization table in the JPEG compression framework leading to dropping information in frequency domain.
Compared with traditional adversarial attack methods (\emph{e.g.} PGD~\cite{madry2018towards}), adversarial examples generated by {AdvDrop} have fewer details, is with decreased image size and possess a higher robustness with respect to denoising-based adversarial defense methods.
However, the confidence of prediction could hardly be improved due to limited and reduced information in the generated adversarial examples. Moreover, there are visible quantization artifacts since the input image is splitted into blocks before transformation.

\textbf{Imperceptibility of adversarial examples} is an important criterion for adversarial attacks, however, it is not well attained by many well-known adversarial attack methods and there usually contains noticeable adversarial perturbations to human-beings. For wider applications, imperceptible adversaries can be applied in privacy protection \textit{e.g.} face recognition. Recently, there are an increasing number of works aiming to improve the imperceptibility of adversarial perturbations~\cite{luo2018towards,croce2019sparse,jia2022exploring,tian2022imperceptible}. Zhao \textit{et al.} \cite{zhao2020towards} introduce PerC-AL in which adversarial perturbations are optimized in terms of perceptual color distance leading to improve visual imperceptibility. Luo \textit{et al.} \cite{luo2022frequency} propose a semantic similarity attack and introduce a new constraint on low-frequency sub-bands between benign images and adversaries, which encourages to add distortions on the high-frequency sub-bands.

\subsection{Invertible Neural Networks}
{Invertible Neural Networks} (INNs)~\cite{dinh2014nice, dinh2016density, kingma2018glow, jacobsen2018revnet} are bijective function approximators due to their mathematically induced network architecture. Given an intermediate feature, INNs are able to explicitly perfect reconstruct features of other layers. That is, the information of INNs' input is preserved throughout all its layers and there is no extra information injected or lost.

INNs are able to explicitly construct inverse mapping, therefore are suitable candidates to perform mappings between two domains. INNs have been applied in many computer vision tasks, including image rescaling~\cite{xiao2020invertible,xiao2022invertible,zhang2022enhancing}, image colorization~\cite{ardizzone2019guided,zhao2021invertible}, video super-resolution~\cite{zhu2019residual,huang2021video}, image denoising~\cite{Liu_2021_CVPR, huang2021winnet, huang2021linn}, image separation~\cite{huang2022durrnet}, image steganography~\cite{xu2021compact,lu2021large,Jing2021HiNetDI,guan2022deepmih}, and invertible image conversion~\cite{cheng2021iicnet}, etc. 
The most relevant to our work is AdvFlow~\cite{mohaghegh2020advflow} which utilizes the normalizing flows to model the density of adversarial examples for black-box adversarial attack.
However, to the best of our knowledge, there is no work using the information preservation property of Invertible Neural Networks for generating adversarial examples.

\section{Proposed Method}
\label{sec:methodology}
Adding bounded class-specific adversarial information to a benign image and dropping existing discriminant information of the original class are two distinctive perspectives for generating adversarial examples and with their own strengths. In this paper, we aim to combine the best features of two paradigms. That is, crafting imperceptible and robust adversarial examples by simultaneously adding and dropping semantic information in an unified framework.

\subsection{Overview}
Given a benign image $\bm{x}_{cln}$ with class label $c$, our objective is to generate a corresponding adversarial image $\bm{x}_{adv}$ by dropping discriminate information of class $c$ while adding adversarial details from a target image $\bm{x}_{tgt}$, and at the same time to be able to interpret the features that have been added or dropped through a residual image $\bm{x}_r$. 

Fig. \ref{fig:overview} shows the overview of the proposed Adversarial attack using Invertible Neural Networks (AdvINN) method. The proposed AdvINN $f_{\bm{\theta}}(\cdot, \cdot)$ is parameterized by $\bm{\theta}$ and with $(\bm{x}_{adv}, \bm{x}_{r}) = f_{\bm{\theta}}(\bm{x}_{cln}, \bm{x}_{tgt})$, where $\bm{\theta}$ represents the parameters of AdvINN. It consists of an Invertible Information Exchange Module (IIEM), a Target Image Learning Module (TILM) and loss functions for optimization. As the source of adversarial information, a target image $\bm{x}_{tgt}$ can be chosen as the highest confidence target image (HCT), an universal adversarial perturbation (UAP), or an online learned classifier guided target image (CGT) using TILM. With $(\bm{x}_{cln}, \bm{x}_{tgt})$, IIEM driven by the loss functions generates the adversarial image $\bm{x}_{adv}$ by performing information exchange between the two images. 
Owing to the information preservation property of IIEM, the amount of information within input images ($\bm{x}_{cln}$, $\bm{x}_{tgt}$) and output images ($\bm{x}_{adv}$, $\bm{x}_{r}$) is the same and there explicitly exists an inverse mapping with $(\bm{x}_{cln}, \bm{x}_{tgt}) = f_{\bm{\theta}}^{-1}(\bm{x}_{adv}, \bm{x}_{r})$.

The learning objective of the proposed AdvINN method can be expressed as:
\begin{equation}
    \begin{aligned}
    \bm{x}_{adv} = &\arg \underset{\bm{\theta}}{\min} \lambda_{adv} \mathcal{L}_{adv}\left(\bm{x}_{adv}, c\right) + \mathcal{L}_{rec}(\bm{x}_{adv}, \bm{x}_{cln}), \\
    & \text {s.t.}\left\|\bm{x}_{adv}-\bm{x}_{cln}\right\|_{\infty} \leq \epsilon,
    \end{aligned}
    \label{eq:obj}
\end{equation}
where $\bm{\theta}$ denotes the parameters of AdvINN, $\mathcal{L}_{adv}(\cdot, \cdot)$ denotes the adversarial loss, $\mathcal{L}_{rec}(\cdot, \cdot)$ denotes the reconstruction loss, $\lambda_{adv}$ is the regularization parameter and $\epsilon$ denotes the budget of adversarial perturbation.

In the following, we will introduce the details of Invertible Information Exchange Module, target image selection and learning, and the loss functions for optimizations.

\subsection{Invertible Information Exchange Module}

To achieve simultaneously adding and dropping semantic information for adversarial example generation, an Invertible Information Exchange Module (IIEM) is proposed as a non-linear transform with information preservation property to interchange information between the clean image and the target image.

\paragraph{Discrete Wavelet Transform.} In order to disentangle the input clean and target images into low-frequency and high-frequency components, Discrete Wavelet Transform (DWT)~\cite{mallat1989theory} has been applied to the inputs for decomposition. The separation of low- and high-frequency features will facilitate modifications to the input image applied on the high-frequency components and therefore results in less perceptible adversarial examples.

With DWT $\mathcal{T}(\cdot)$, an input image $\bm{x} \in \mathbb{R}^{C\times H \times W}$ will transformed into wavelet domain $\mathcal{T}(\bm{x}) \in \mathbb{R}^{4C\times H/2 \times W/2}$. 
It contains one low-frequency sub-band feature and three high-frequency sub-band features.
At the output end of IIEM, Inverse Discrete Wavelet Transform (IDWT) $\mathcal{T}^{-1}(\cdot)$ has been applied to reconstruct the features back to image domain. 

\paragraph{Affine Coupling Blocks.}
Invertible Information Exchange Module is composed of $M$ Affine Coupling Blocks. Let us denote with $\bm{w}_{cln}^{i}$ and $\bm{w}_{tgt}^{i}$ the input features of the $i$-th Affine Coupling Block, and with $\bm{w}_{cln}^{0}=\mathcal{T}(\bm{x}_{cln})$ and $\bm{w}_{tgt}^{0}=\mathcal{T}(\bm{x}_{tgt})$. Then, the forward process of the $i$-th Affine Coupling Block can be expressed as:
\begin{equation}
    \begin{aligned}
        \bm{w}_{cln}^{i} = &\bm{w}_{cln}^{i-1} \odot \exp \left(\alpha\left({\psi}\left(\bm{w}_{tgt}^{i-1}\right)\right)\right)+{\phi}\left(\bm{w}_{tgt}^{i-1}\right),\\
        \bm{w}_{tgt}^{i} = &\bm{w}_{tgt}^{i-1} \odot \exp \left(\alpha\left({\rho}\left(\bm{w}_{cln}^{i}\right)\right)\right)+{\eta}\left(\bm{w}_{cln}^{i}\right),
    \end{aligned}
    \label{eq:ACB}
\end{equation}
where $\odot$ denotes element-wise multiplication, $\bm{\alpha}$ is a Sigmoid function multiplied by a constant factor, and $\psi(\cdot), \phi(\cdot), \rho(\cdot), \eta(\cdot)$ denote dense network architectures as in \cite{wang2018esrgan}.

Given the output of $M$-th Affine Coupling Block, the adversarial image and the residual image can be reconstructed using IDWT with $\bm{x}_{adv} = \mathcal{T}^{-1}(\bm{w}_{cln}^{M})$ and $\bm{x}_{r} = \mathcal{T}^{-1}(\bm{w}_{tgt}^{M})$.

By default, for DWT/IDWT, Haar wavelet transform is used, and the number of Affine Coupling Blocks is set to 2.

\paragraph{Information Preservation Property.}
Due to the invertibility of DWT and IDWT, $\bm{w}_{cln}^{M}$ and $\bm{w}_{tgt}^{M}$ can be restored from $\left( \bm{x}_{adv}, \bm{x}_{r} \right)$.
In IIEM, only the forward process of the Affine Coupling Blocks is used for generating adversarial images, and it's worth to note that $(\bm{w}_{cln}^{i-1}, \bm{w}_{tgt}^{i-1})$ can be perfectly recovered from $(\bm{w}_{cln}^{i}, \bm{w}_{tgt}^{i})$:
\begin{equation}
    \begin{aligned}
        \bm{w}_{tgt}^{i-1} =& \left(\bm{w}_{tgt}^{i} - {\eta}\left(\bm{w}_{cln}^{i}\right) \right) \odot \exp \left(-\alpha({\rho}(\bm{w}_{cln}^{i}))\right),\\
        \bm{w}_{cln}^{i-1} = & \left(\bm{w}_{cln}^{i} - {\phi}\left(\bm{w}_{tgt}^{i-1}\right) \right)\odot \exp \left(-\alpha({\psi}(\bm{w}_{tgt}^{i-1}))\right).
    \end{aligned}
    \label{eq:ACBinv}
\end{equation}

Therefore, IIEM is fully invertible and the output images $(\bm{x}_{adv}, \bm{x}_{r})$ contain the same amount of information as the input images $(\bm{x}_{cln}, \bm{x}_{tgt})$.
Their relationship can be represented as:
\begin{equation}
    \begin{cases}
        \bm{x}_{adv} &= \bm{x}_{cln} - \bm{\sigma} + \bm{\delta}, \\
        \bm{x}_{r} &= \bm{x}_{tgt} + \bm{\sigma} - \bm{\delta}.
    \end{cases}
\end{equation}
where $\bm{\sigma}$ denotes the dropped existing information of the clean image, and $\bm{\delta}$ denotes the added discriminant information from the target image to the clean image.

%-----------------------------------------------------------
\begin{table*}[t]
  \centering
  \caption{Accuracy and evaluation metrics on different methods. All methods use $\epsilon =8/255$ as the adversarial budget. ASR donates the accuracy of adversarial attacks. $\uparrow$ means the value is higher the better, and vice versa. (The best and the second best result in each column is in bold and underline.)}
    \begin{tabular}{l|l|C{1.5cm}C{1.5cm}C{1.5cm}C{1.5cm}C{1.5cm}C{1.5cm}C{1.5cm}}
    \toprule
    Dataset & Methods &  $l_{2}\downarrow$ & $l_{\infty}\downarrow$& SSIM$\uparrow$    & LPIPS$\downarrow$ & FID$\downarrow$   & ASR(\%)$\uparrow$ \\
    \midrule
    \midrule
    \multirow{9}{*}{ImageNet-1K} 
        & StepLL   & 26.90 &0.04 & 0.948 & 0.1443 & 25.176 & 98.5 \\
        & C\&W   & 10.33 & 0.07 & 0.977 & 0.0617 & 11.515 & 91.7 \\
        & PGD    & 64.42 & 0.04 & 0.881 & 0.2155 & 35.012 & 90.2 \\
        & PerC-AL    & \textbf{1.93} & 0.10 & \underline{0.995} & {0.0339} & {5.118} & \textbf{100.0} \\
        & AdvDrop  & 18.47 & 0.07 & 0.977 & 0.0639 & 9.687 & \textbf{100.0} \\
        & SSAH    & 6.97 & \textbf{0.03} & 0.991 & 0.0352 & 5.221 & \underline{99.8} \\
        & AdvINN-HCT & 5.73 & \textbf{0.03} & 0.991 & \underline{0.0206} & 3.661 & \textbf{100.0}\\
        & AdvINN-UAP & 5.84 & \textbf{0.03} & 0.990 & {0.0212} & \underline{2.900} & \textbf{100.0}\\
        & AdvINN-CGT & \underline{2.66} & \textbf{0.03} & \textbf{0.996} & \textbf{0.0118} & \textbf{1.594} &\textbf{100.0}\\
    \midrule
    \midrule
    \multirow{9}{*}{CIFAR-100} 
        & StepLL  & 0.73  & 0.04 & 0.923 & 0.0411 & 11.608   & 94.3 \\%updated
        & C\&W     & 1.24  & 0.09 & 0.943 & 0.0706 & 12.507  & 97.7 \\%updated
        & PGD    & 1.59  & \textbf{0.03}  & 0.954 & 0.0793 & 23.899 & 99.2 \\%updated
        & PerC-AL   & 3.09  & 0.27 & 0.961 & 0.0426 &  6.035  & 97.2 \\%updated
        & AdvDrop   &  87.09  &  0.61 & 0.774 & 0.2549  & 14.722 & 90.7 \\%updated
        & SSAH   & 0.43  & 0.04 & 0.992 & 0.0200 & 4.508 & 99.4 \\%自己的pt %untarget updated
        & AdvINN-HCT & 0.28 & \textbf{0.03} & \underline{0.991} & \textbf{0.0035} & \textbf{3.413} & 98.3\\%1DWT 8/255
        & AdvINN-UAP & \underline{0.27} & \textbf{0.03} & \textbf{0.993} & \underline{0.0037} & 3.982 & \textbf{99.6}\\%2DWT
        & AdvINN-CGT    &\textbf{0.23}  & \textbf{0.03} & \textbf{0.993} &\underline{0.0037}  & \underline{3.921} & \underline{99.5}  \\%updated_w=0.25
    \midrule
    \midrule
    \multirow{9}{*}{CIFAR-10} 
        & StepLL   & 0.77 & 0.04 & 0.982 & 0.0462 & 10.997 & 98.2 \\%updated
        & C\&W    & 1.06 & 0.09 & 0.970 & 0.0667 & 10.510 & 99.3 \\%updated
        & PGD    & 1.61 & \textbf{0.03} & 0.956 &  0.0861  & 24.014 & \textbf{100.0} \\%updated
        & PerC-AL   & 0.52 & 0.13 & 0.990 & 0.0134 & \textbf{1.518} & \textbf{100.0} \\%updated with 85% confidence
        & AdvDrop   & 70.10 & 0.46 & 0.570 & 0.4483 & 122.950 & 97.7  \\ %updated
        & SSAH   & 0.38 & \textbf{0.03} & \underline{0.993} & 0.0180 & 3.654 & \underline{99.9} \\%updated
        & AvdINN-HCT & \underline{0.18} & \textbf{0.03} & \textbf{0.995} & 0.0033 & 2.627 & \underline{99.9}\\%2DWT 1layer w=0.2
        & AdvINN-UAP & 0.19 & \textbf{0.03} & \textbf{0.995} & \underline{0.0031} & 2.791 & \underline{99.9}\\%2DWT 1layer w=0.2
        & AdvINN-CGT   & \textbf{0.17} & \textbf{0.03} & \textbf{0.995} & \textbf{0.0030} & \underline{2.480} & \underline{99.9} \\%2DWT 1layer w=0.2
    \bottomrule
    \end{tabular}%
  \label{tab:table2}%
\end{table*}%
%-----------------------------------------------------------

In the case of $\bm{x}_{tgt}$ being a constant image, there will be no information added from the target image to the clean image, \textit{i.e.}, $\bm{\delta}=0$. The residual image $\bm{x}_{r}$ will then only correspond to the dropped information $\bm{\sigma}$ and can be used to interpret the results of AdvINN.

\subsection{Target Image Selection and Learning}
\label{sec:Target Image}
The target image in the proposed AdvINN method plays an essential role and determines the information to be added to the clean image for generating the adversarial image. 
In this section, we introduce three options for selecting or learning the target image.

\paragraph{Highest Confidence Target Image (HCT).} 
The most intuitive idea is to select the image with the {{highest confidence}} in each class as the target image as StepLL~\cite{kurakin2016adversarial}, since the higher confidence of the target image to the classifier is, the more discriminant information the images may possess. 
However, selecting a natural image as the target image may not be the best option, since a natural image often carries a considerable amount of information unrelated to the target class, such as background texture and the details of other classes.
This could hinder the optimization speed as well as the success rates of attacks.

\paragraph{UAP as Target Image (UAP).} 
{{Universal Adversarial Perturbations}}~\cite{UAPs2017, poursaeed2018generative,Khrulkov2018ArtOS,zhang2020understanding,benz2020double} aggregate dominant information of images of the target class and minimize the interference of irrelevant details. Therefore, it could be better option for the target image.
Zhao \emph{et al.}\cite{NEURIPS2021_30d454f0} propose that targeted adversarial perturbations are optimized in a data-free manner.
We follow this method and utilize the optimized universal adversarial perturbation as target images, which is able to moderately accelerate convergence speed.

\paragraph{Classifier Guided Target Image (CGT).} 
In order to further improve the optimization speed as well as attacking success rate, we propose a Target Image Learning Module (TILM) which learns a classifier guided target image rather than using a fixed image as the target image.
Inspired by targeted UAPs~\cite{NEURIPS2021_30d454f0}, the target image is set to be a learnable variable which is initialized with a constant image (\textit{i.e.,} all pixels are set to 0.5) and then updated according to the gradient from the attacking classifier. In this way, an adaptively generated target image can embed more discriminant information of the target class assisting the generation of adversarial examples. 
Detailed experimental results on the target image selection will be discussed and presented in Experiments section.

\subsection{Learning Details}
\label{sec:loss function}
We set up a reconstruction loss $\mathcal{L}_{{rec}}$ and an adversarial loss $\mathcal{L}_{{adv}}$ to locate the correct optimized direction and accelerate convergence speed. The total loss can be expressed as:
\begin{equation}
    % \label{equation9}
    \mathcal{L}_{{total}}= \lambda_{adv} \mathcal{L}_{{adv}} + \mathcal{L}_{{rec}},
    \label{eq:totalloss}
\end{equation}
where $\lambda_{adv}$ is the regularization parameter.

\noindent \textbf{Reconstruction loss} is utilized to constrain the optimized adversarial image being close to the input clean image, while imposing the modifications to be applied mainly on the high-frequency and less perceptible components leading to less visible adversarial examples:
\begin{equation}
    \begin{aligned}
        \label{equation5}
        \mathcal{L}_{{rec}}=&\sum_{i\in \{ll, lh, hl, hh \}} w_i \Vert \mathcal{T}(\bm{x}_{{cln}})_i, \mathcal{T}(\bm{x}_{{adv}})_i \Vert_2^2\\
        &+ \lambda_{perp}  \Vert \rho(\bm{x}_{{cln}}), \rho(\bm{x}_{{adv}}) \Vert_2^2,
    \end{aligned}
\end{equation}
where ${ll, lh, hl, hh}$ denote the low- and high-frequency components of the wavelet transform, $ w_i$ is the weight of the corresponding wavelet component, $\lambda_{perp}$ is the weight of perceptual loss and $\rho(\cdot)$ denotes the features of the VGG-16 model pretrained on ImageNet dataset.

\noindent \textbf{Adversarial loss} evaluates dissimilarity of prediction logits and target label. The cross entropy loss $\ell_{CE}(\cdot)$ is used to measure the difference.
\begin{equation}
    \label{equation8}
    \mathcal{L}_{{adv}}=\ell_{CE}\left(g_{\phi}({\bm{x}_{{adv}}}), {c}_{{tgt}}\right),
\end{equation} 
where $g_{\phi}(\cdot)$ denotes the target classifier, and ${c}_{tgt}$ is the label of the target class.

We set a classifier guided loss $\mathcal{L}_{{cgt}}$ for learning $\bm{x}_{cgt}$. It is also a cross entropy loss similar to (\ref{equation8}).

%-----------------------------------------------------------
\begin{figure*}[t]
    \centering
    \subfigure[Clean]{  
    \includegraphics[width=0.18\textwidth]{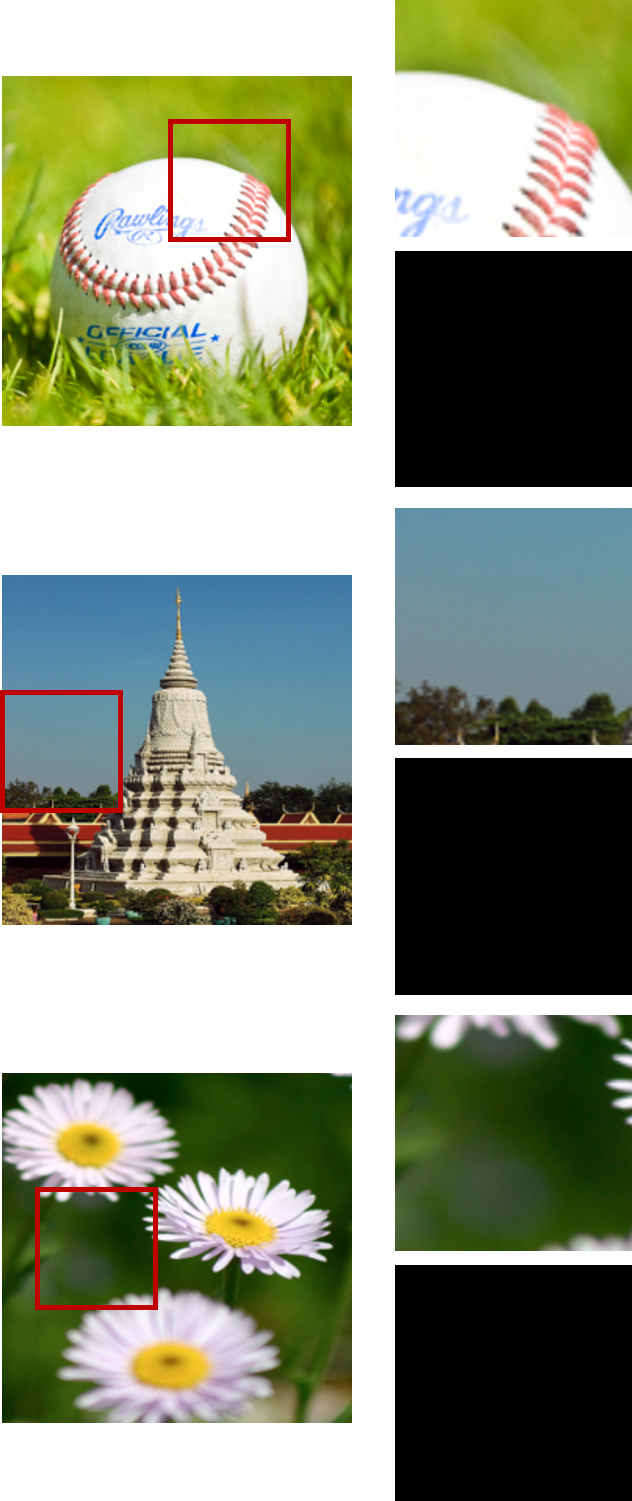}
    }
    \subfigure[C\&W]{
    \includegraphics[width=0.18\textwidth]{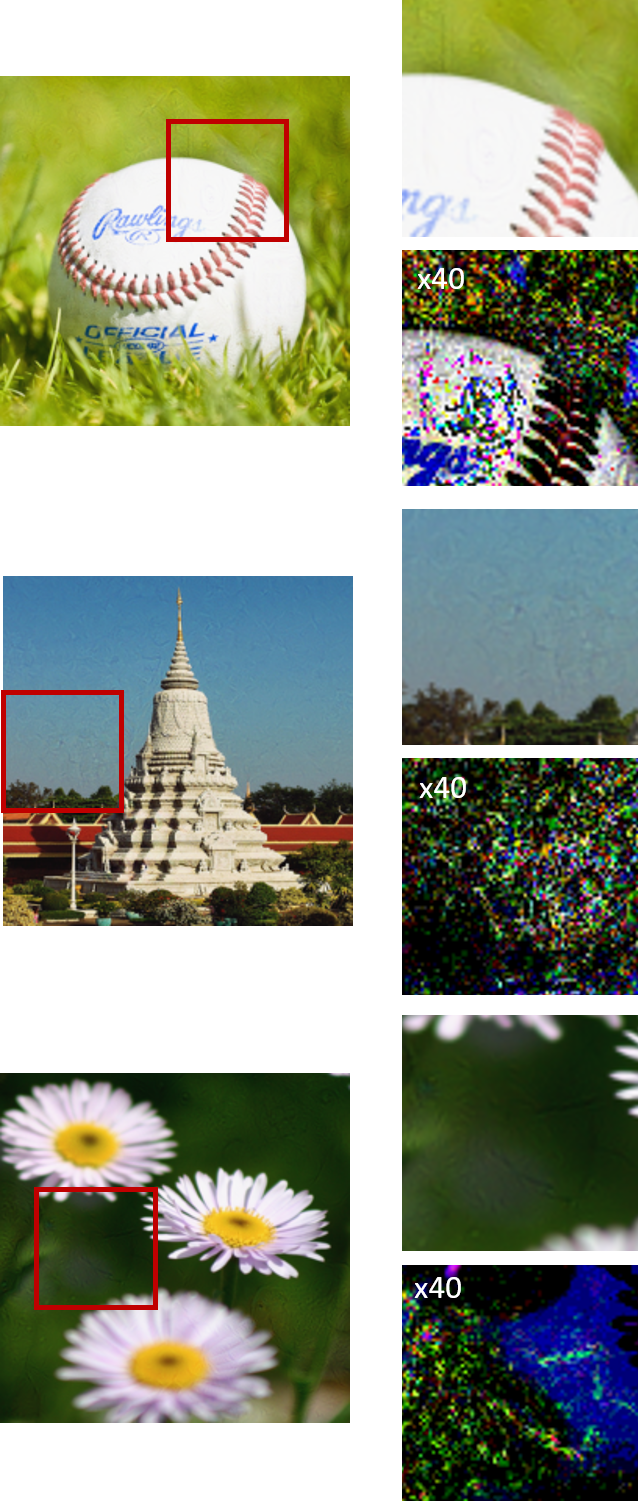}
    }
    \subfigure[SSAH]{
    \includegraphics[width=0.18\textwidth]{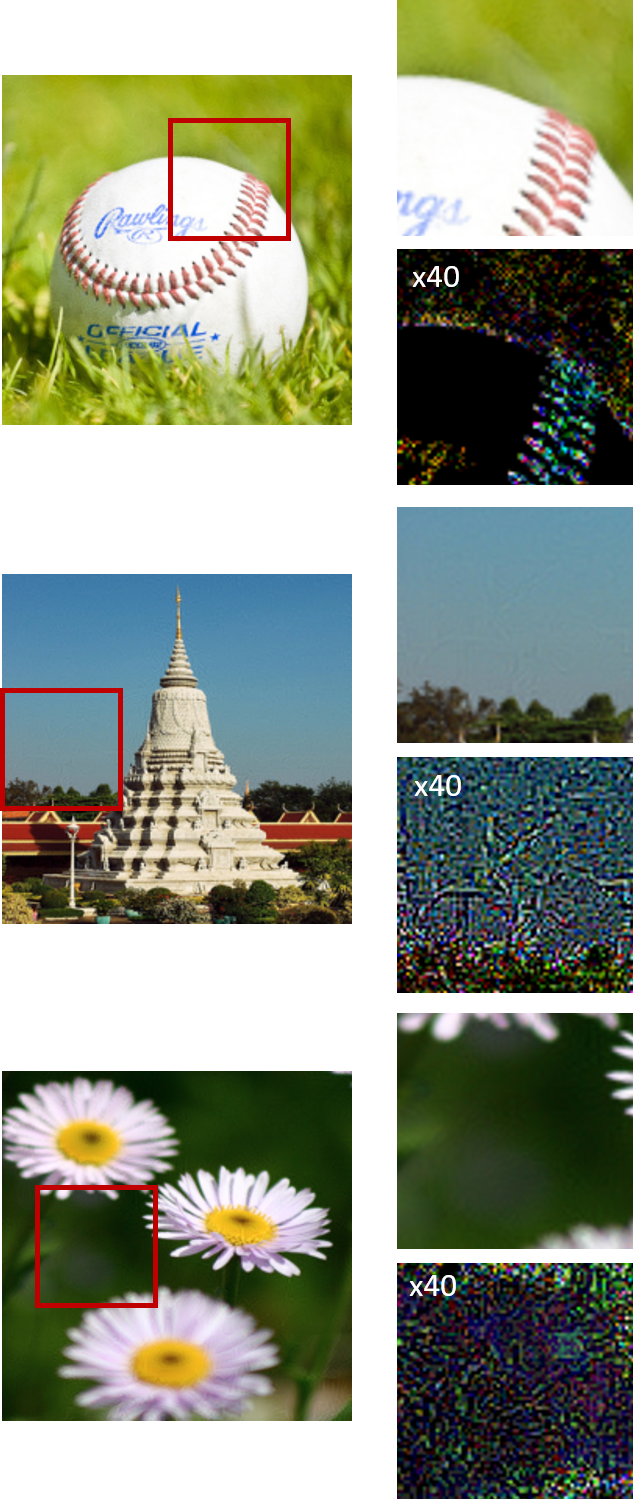}
    }
    \subfigure[PerC-AL]{
    \includegraphics[width=0.18\textwidth]{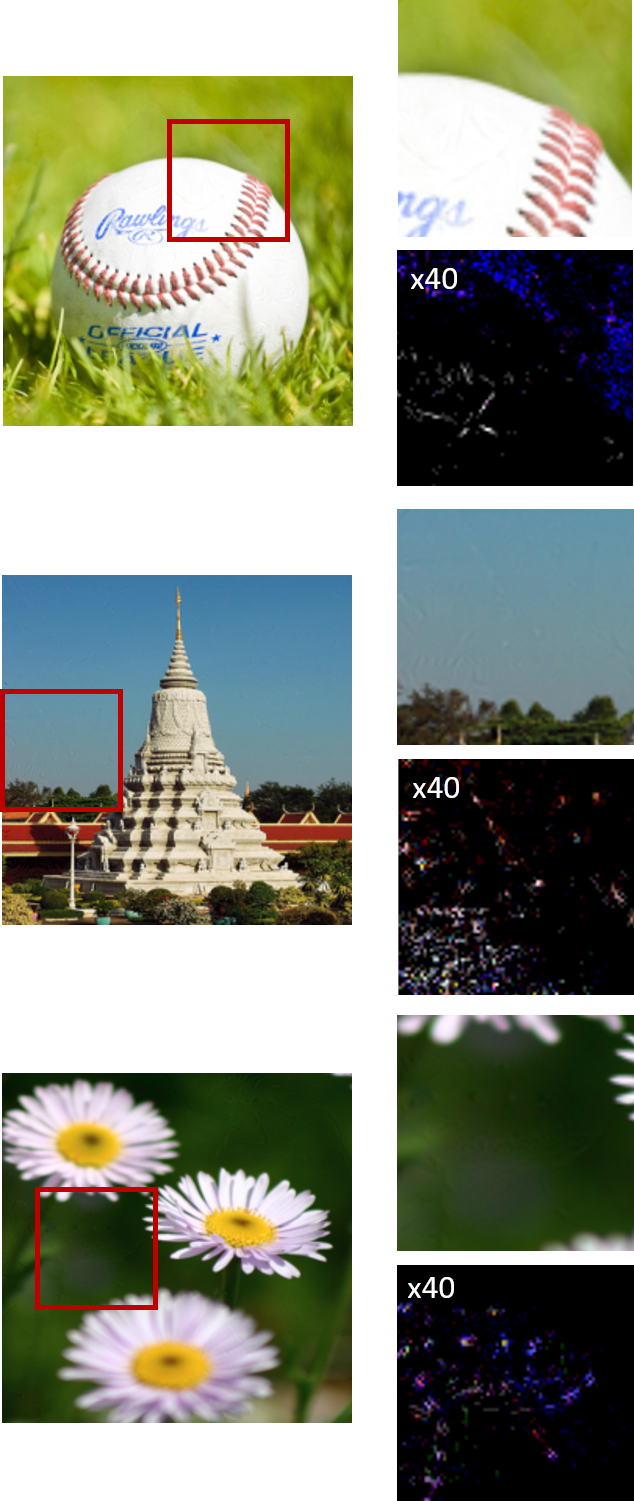}
    }
    \subfigure[AdvINN-CGT]{
    \includegraphics[width=0.18\textwidth]{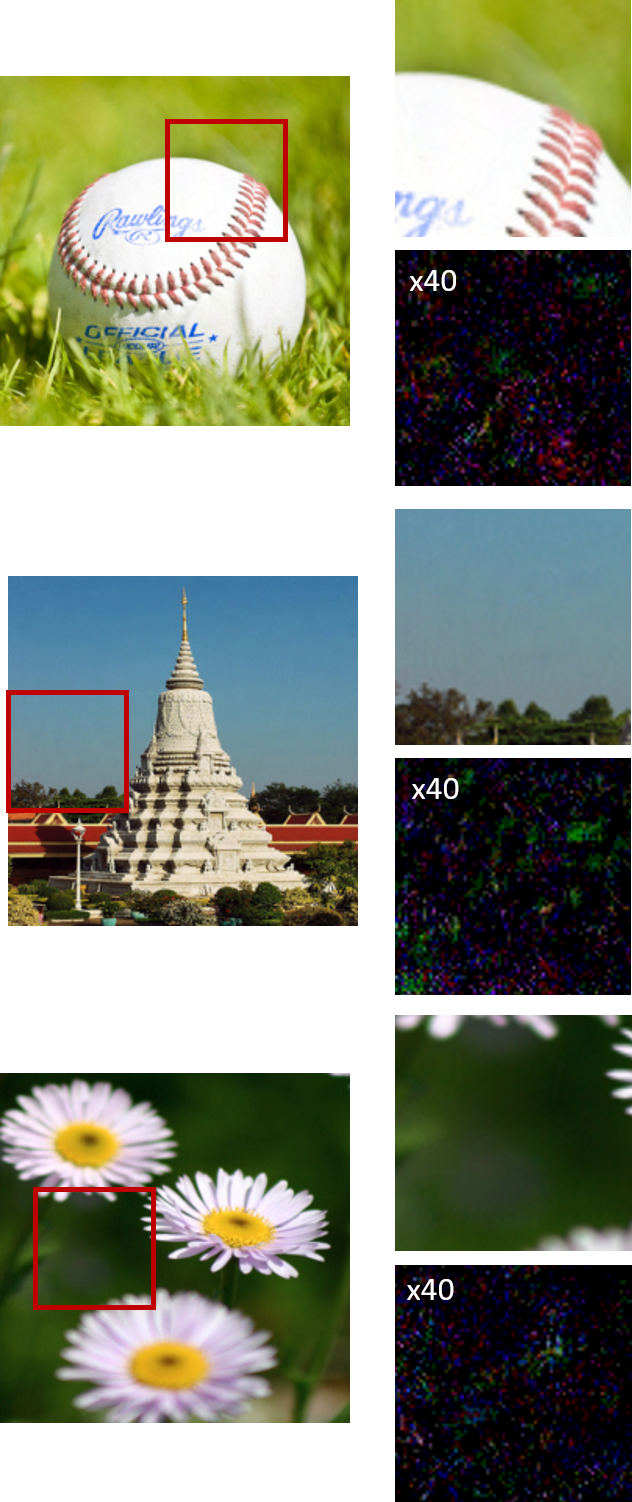}
    }
    \caption{Exemplar adversarial examples and the corresponding adversarial perturbations generated by C\&W, SSAH, PerC-AL, and the proposed AdvINN method on ImageNet-1K.}
    \label{fig:imagenet}
\end{figure*}
%-----------------------------------------------------------

The optimizer for optimizing the learning objective of AdvINN in (\ref{eq:obj}) is set to Adam~\cite{kingma2014adam} optimizer with initial learning rate $1e^{-4}$ which is decayed every 200 iterations with decay rate 0.9 and is lower bounded by $1e^{-5}$. We empirically set the regularization parameters $\lambda_{adv}=3$, $w_{ll}=2$, $w_{lh,hl,hh}=1$ and $\lambda_{perp}=0.001$. 
% The code of the proposed method will be publicly available.

\section{Experiments}
\label{sec:experiments}
\subsection{Experimental Setup} 
\noindent \textbf{Dataset and models.}
We evaluate the performance of the comparison methods on ImageNet-1K dataset which contains 1000 images
% \footnote{The images are resized to $256 \times 256$ and then center cropped to $224 \times 224$. } 
sampled from the ImageNet-1K validation set~\cite{russakovsky2015imagenet}.  The benign images are all correctly classified by the target classifier.
% More experimental results on CIFAR-10 and CIFAR-100 are shown in the supplementary material. 
All experiments were performed on a computer with a NVIDIA RTX 3090 GPU with 24 GB memory.

\noindent \textbf{Comparison Methods.}
Six comparison methods have been included for evaluation, with three well-known adversarial attack methods as our baselines including PGD~\cite{madry2018towards} under $l_{\infty}$-norm, StepLL~\cite{kurakin2016adversarial}, and C\&W~\cite{carlini2017towards}, and three recent state-of-the-art methods, including AdvDrop~\cite{duan2021advdrop}, PerC-AL~\cite{zhao2020towards} and SSAH~\cite{luo2022frequency}.

\noindent \textbf{Evaluation metrics.}
We use attacking success rate (ASR) to evaluate the attacking performance, and five popular metrics to evaluate the quality of the generated adversarial images, including: $l_{2}$-norm, $l_{\infty}$-norm, Structural Similarity Index (SSIM)~\cite{wang2004image}, Learned Perceptual Image Patch Similarity (LPIPS)~\cite{zhang2018unreasonable} and Fréchet Inception Distance (FID)~\cite{heusel2017gans}. Specifically, $l_{2}$-norm measures the average energy of the adversarial perturbations, $l_{\infty}$-norm evaluates the maximum perturbation intensity, SSIM assesses the structural similarity between two images, and LPIPS and FID both measure the perceptual similarity.

\noindent \textbf{Attack setting.}
For fair comparisons, all comparison methods perform targeted attacks with the least-likely objective (except SSAH)
% \footnote{Restricted by the memory, SSAH chooses the most dissimilar image as target in a batch of size 200.}
 to avoid choosing closely related classes which is less meaningful in real applications. 
For the target classifier, we use pre-trained ResNet50\footnote{https://download.pytorch.org/models} which is with 23.85\% top-1 error on ImageNet-1K. 
% Experimental results on Inceptionv3~\cite{szegedy2016rethinking} and Densenet121~\cite{huang2017densely} are shown in the supplementary material.

\subsection{Evaluation on Targeted Attacks}
\label{sec:evaluation}

%-----------------------------------------------------------
\begin{figure}[t]
    \centering
    \includegraphics[width=1.0\linewidth]{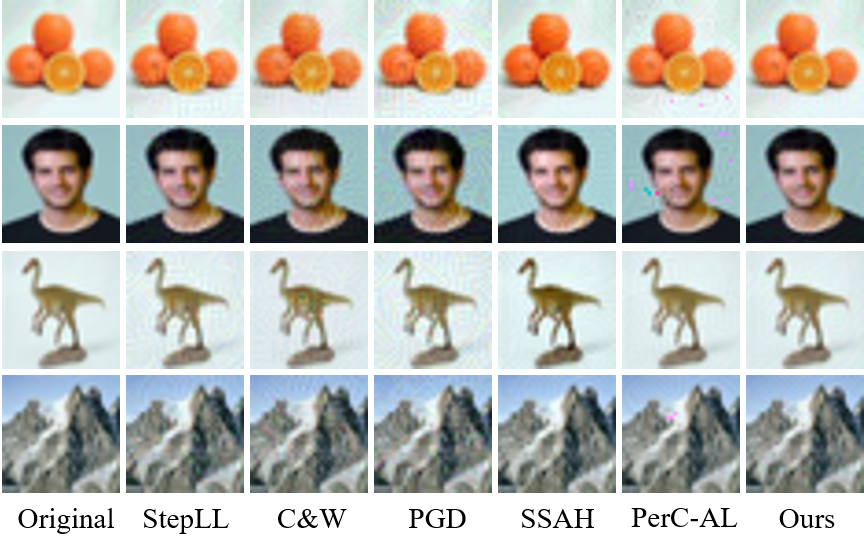}
    \caption{More adversarial examples crafted by different methods on CIFAR-100.}
    \label{fig:cifar100}
\end{figure}
%-----------------------------------------------------------
%-----------------------------------------------------------
\begin{figure}[t]
    \centering
    \includegraphics[width=1.0\linewidth]{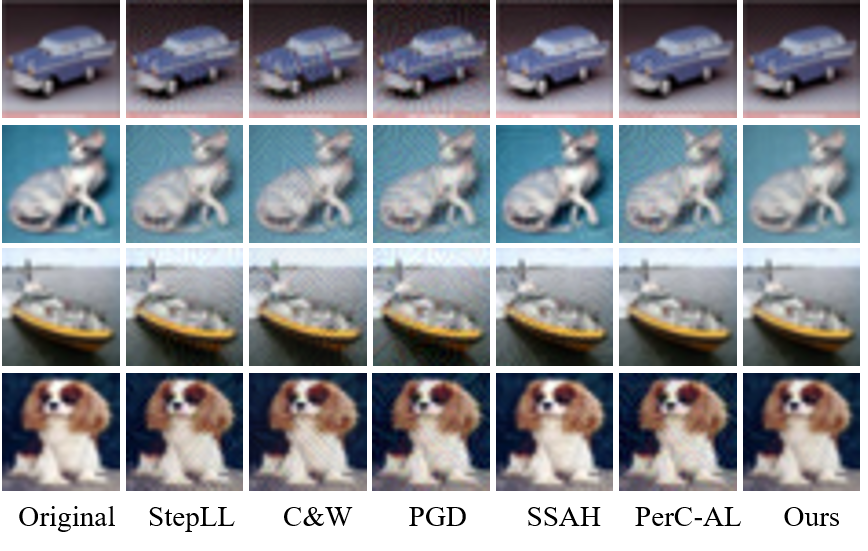}
    \caption{More adversarial examples crafted by different methods on CIFAR-10.}
    \label{fig:cifar10}
\end{figure}
%-----------------------------------------------------------

Table \ref{tab:table2} shows the white-box targeted attack performance of different methods on ImageNet-1K as well as the quality of the adversarial images evaluated using $l_{2}$-norm, $l_{\infty}$-norm, SSIM, LPIPS, and FID. We can see that under the same perturbation budget $\epsilon=8/255$  with respect to $l_{\infty}$-norm, the proposed AdvINN method with HCT, UAP, and CGT achieve better image quality, especially on perceptual metrics, than the state-of-the-art methods. Specifically, AdvINN-CGT achieves the best results in terms of SSIM, LPIPS and FID and is with 100\% ASR.
In terms of FID, AdvINN-CGT achieves 8.093, 3.627, 3.524 lower FID score compared to AdvDrop~\cite{duan2021advdrop}, SSAH~\cite{luo2022frequency}, PerC-AL~\cite{zhao2020towards}, respectively.
PerC-AL achieves lower $l_{2}$-norm but the largest $l_{\infty}$-norm, because it modifies a smaller number of pixels but with significantly larger values among all comparison methods. Therefore, PerC-AL has unsatisfactory perceptual scores. AdvINN-CGT achieves the 2$^{nd}$ lowest $l_{2}$-norm, and the best scores in all other metrics.
This indicates that the adversarial examples generated by AdvINN have higher structural and perceptual similarity to the ground-truth images than those of comparison methods.

Fig. \ref{fig:imagenet} shows the exemplar adversarial examples crafted by different methods on ImageNet-1K. We can see that the proposed AdvINN method is able to generate visually less perceptible adversarial examples than comparison methods. The possible reason is that AdvINN performs information exchange on feature domain via Invertible Neural Networks. 
While traditional methods that generate adversarial samples based on the image domain usually restrict the number of perturbations by $l_{2}$-norm and $l_{\infty}$-norm leading to uniformly distributed adversarial perturbations over the whole image.

We have also evaluated the performance of all comparison methods on the testing set of CIFAR-10 and CIFAR-100. The benign images are all correctly classified by the target classifier. For the target classifier, we use pre-trained ResNet-20\footnote{https://github.com/chenyaofo/pytorch-cifar-models} with 7.4\% and 30.4\% top-1 error on CIFAR-10 and CIFAR-100, respectively.

The parameter setting of AdvINN for CIFAR-10 and CIFAR-100 is slightly different from those used in ImageNet-1K due to the large difference on testing image size. 
The optimal setting for AdvINN is to use 1 Affine Coupling Block and 2 DWT layers. And the hyper-parameter $\mathcal{\lambda}_{adv}$ is set to 0.25 for CIFAR-100 and 0.2 for CIFAR-10 in order to craft more imperceptible adversarial examples.

Table \ref{tab:table2} shows the white-box targeted attack performance of different methods on CIFAR-100 and CIFAR-10 as well as the quality of the adversarial images evaluated using $l_{2}$-norm, $l_{\infty}$-norm, SSIM, LPIPS, and FID. We can observe that the proposed AdvINN method achieves less perceptible adversarial examples and guarantees high attacking success rate. 
Fig. \ref{fig:cifar100} and Fig. \ref{fig:cifar10} show the adversarial examples generated on CIFAR-100 and CIFAR-10, which can also support that our method generates more imperceptible examples in human visual system than other methods.

Table \ref{tab:table2} also depicts the performance of AdvINN with different choices of target images. With highest confidence target images, AdvINN-HCT produces satisfactory image quality, however natural images usually contain irrelevant details to the target class that may slow down the optimization speed. By utilizing UAPs which effectively excludes unrelated information, AdvINN-UAP achieves around 20\% speed-up in optimization speed and is with similar performance to AdvINN-HCT. 
However, the highest confidence images and targeted UAPs both need to prepare beforehand and may not contain suitable features to be transferred. 
AdvINN-CGT instead uses a learnable target image which is generated through the guidance of the classifier and learns essential details to the objective. We can see that AdvINN-CGT achieves the best image quality, and moreover, it only takes around 10\% iterations to converge compared to AdvINN-UAP. Unless otherwise specified, we refer AdvINN as AdvINN-CGT.

%-----------------------------------------------------------
\begin{figure*}[!h]
    \centering
    \subfigure[JPEG Compression]{
    \includegraphics[width=0.3\textwidth]{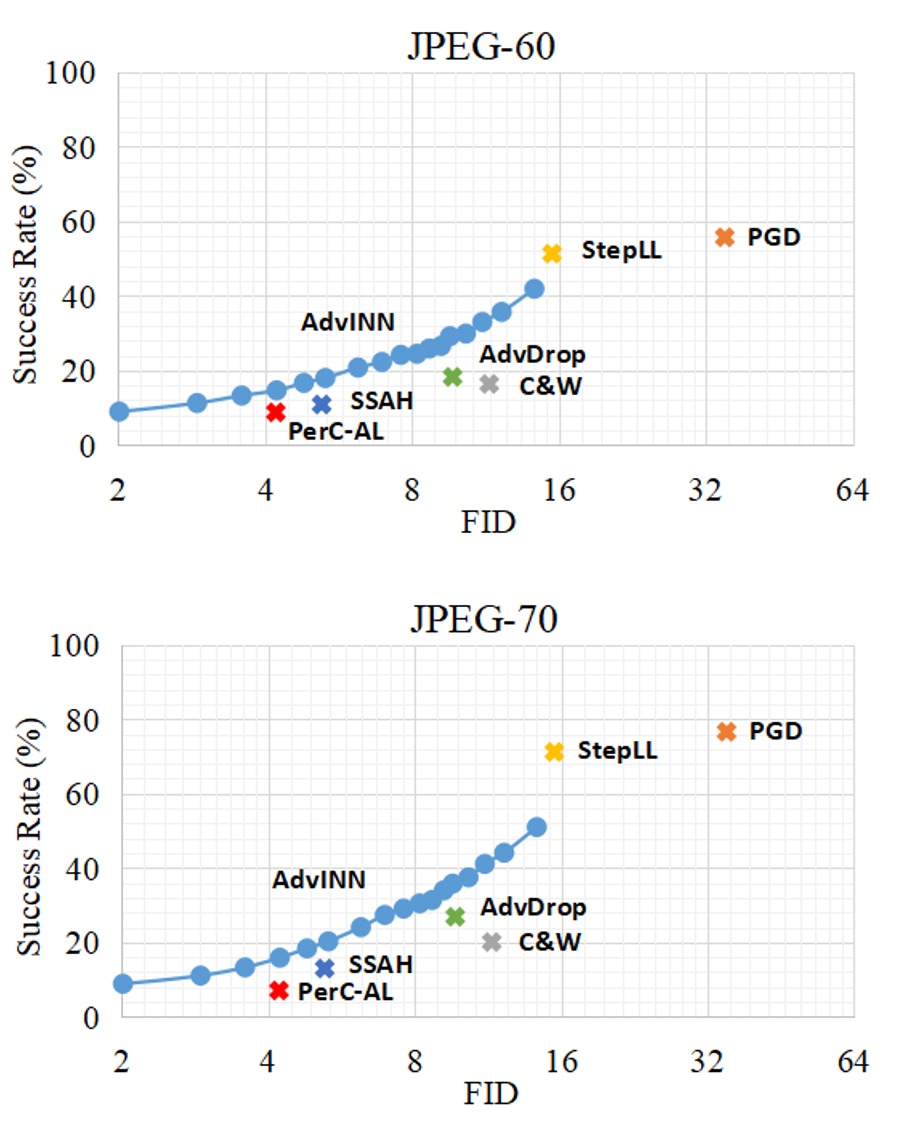}
    }
    \subfigure[Bit-Depth Reduction]{
    \includegraphics[width=0.305\textwidth]{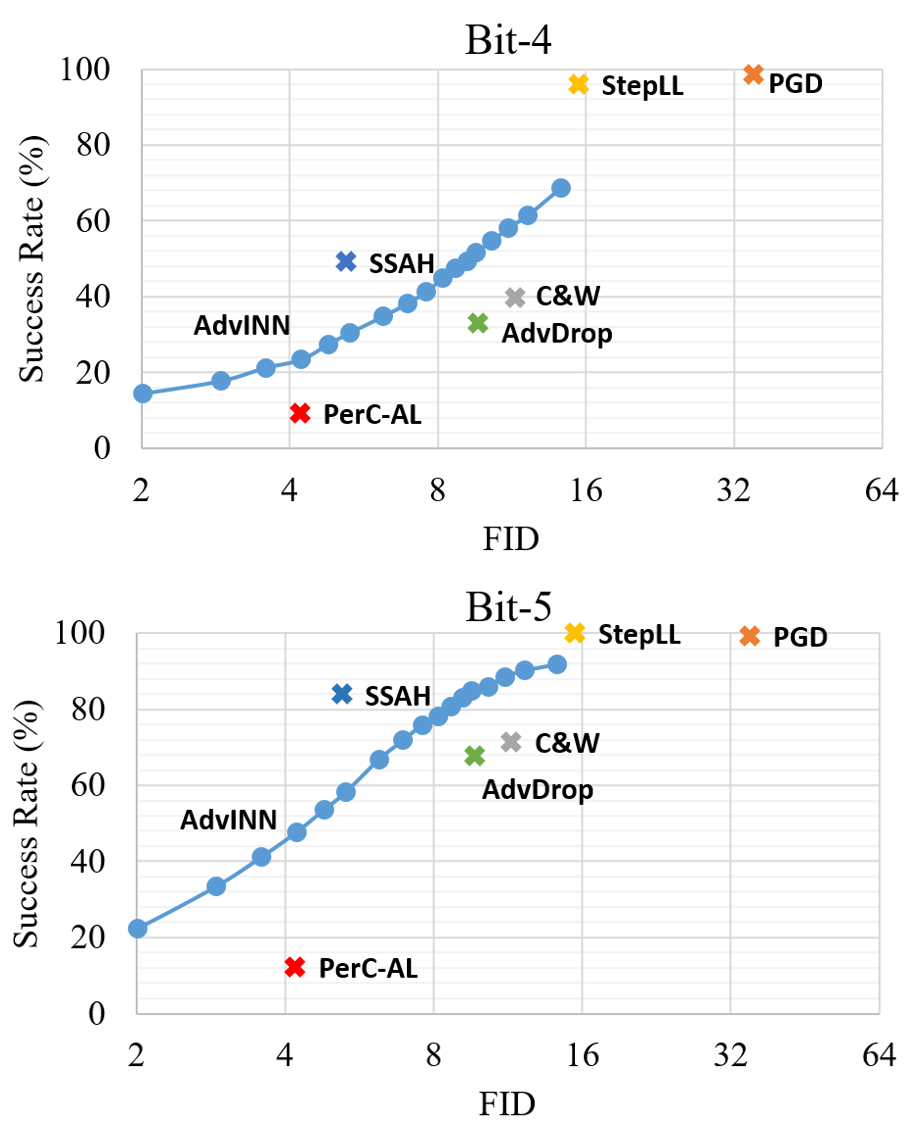}
    }    
    \subfigure[NRP]{
    \includegraphics[width=0.298\textwidth]{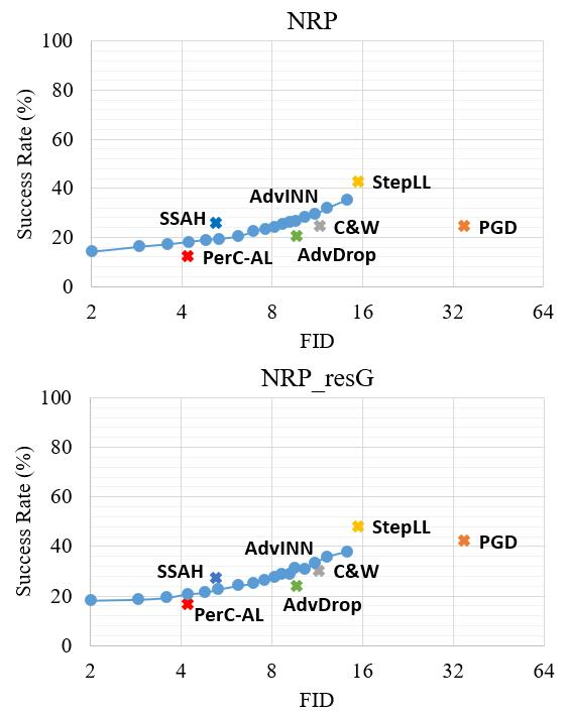}
    }
    \caption{Evaluation on robustness of adversarial examples. All other methods use its recommended parameter settings.}
    \label{fig:defense2}
\end{figure*}
%-----------------------------------------------------------
\subsection{Robustness}
We follow robustness evaluation settings in AdvDrop~\cite{duan2021advdrop} and PerC-AL~\cite{zhao2020towards} and choose two common defense methods based on image transformation, \textit{i.e.}, JPEG compression~\cite{2018Shield} and bit-depth reduction~\cite{2017Countering}. 
Except that, we have added purification-based methods named NRP and NRP\_resG~\cite{naseer2020self} to investigate the robustness of the adversarial examples.
Fig. \ref{fig:defense2} illustrates the imperceptibility and robustness of the adversarial examples generated by different methods. Specifically, the horizontal and vertical axis represents the FID score and the attacking success rate, respectively. 

By varying the regularization parameter ${\lambda_{adv}}$ in (\ref{eq:totalloss}) within the range of $[3, 400]$, the proposed AdvINN method can achieve a trade-off between model robustness and imperceptibility and at the same time ensure 100\% attacking success rate.
% (more specific results are shown in the supplementary material).
We can see that the adversarial examples generated by PGD and StepLL are more robust against defense methods, however, the significantly higher FID scores indicate that these adversarial examples are too visually perceptible to fool human-beings. 
When FID score is in the range of [2,16], AdvINN outperforms all comparison methods against JPEG compression, and outperforms most comparison methods against bit-depth reduction and NRP  except SSAH. With the same FID constraint, AdvINN can generate more robust adversarial examples, and with the same attacking success rate, AdvINN can achieve a lower visual perceptibility.

\subsection{Effectiveness of IIEM} 

\begin{table}[t]
  \centering
  \caption{Ablation study on the effectiveness of IIEM. Iter represents the average iterations on the whole dataset.}
    \begin{tabular}{l|ccccc}
    \toprule
          Generator & SSIM$\uparrow$  %& PSNR$\uparrow$  
          & $l_{2}\downarrow$    & LPIPS$\downarrow$ & FID$\downarrow$   & Iter$\downarrow$ \\
    \midrule
    \midrule
    w/ IIEM & 0.996 & 2.66 &0.0118 &1.594 & 321 \\
    w/o IIEM & 0.973  & 39.61 & 0.0325 & 6.141 & 272 \\%& 37.10
    w/ CNN &0.901 &46.04 &0.0360 &4.345 &1800 \\
    \bottomrule
    \end{tabular}%
  \label{tab:effectiveness of IIEM}%
\end{table}%

% -----------------------------------------------------------------------------
\begin{figure}[t]
    \centering
    \subfigure[HCT]{
    \begin{minipage}[b]{0.12\textwidth}
    \includegraphics[width=1.0\linewidth]{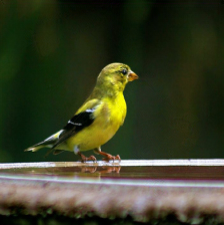}
    \includegraphics[width=1.0\linewidth]{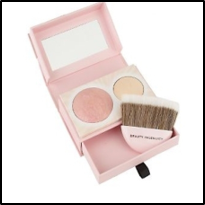}
    \includegraphics[width=1.0\linewidth]{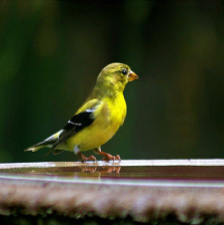}
    \includegraphics[width=1.0\linewidth]{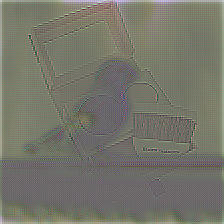}
    \includegraphics[width=1.0\linewidth]{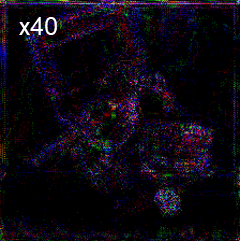}
    \includegraphics[width=1.0\linewidth]{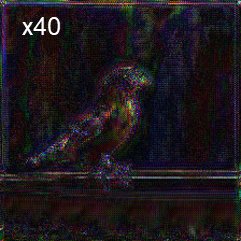}
    \end{minipage}
    }
    \subfigure[UAP]{
    \begin{minipage}[b]{0.12\textwidth}
    \includegraphics[width=1.0\linewidth]{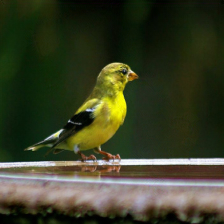}
    \includegraphics[width=1.0\linewidth]{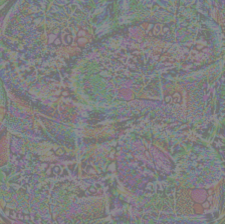}
    \includegraphics[width=1.0\linewidth]{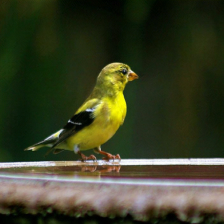}
    \includegraphics[width=1.0\linewidth]{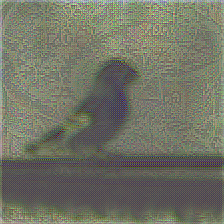}
    \includegraphics[width=1.0\linewidth]{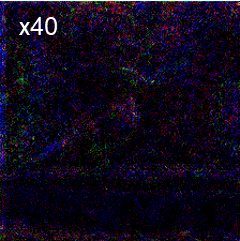}
    \includegraphics[width=1.0\linewidth]{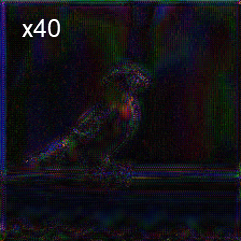}
    \end{minipage}
    }
    \subfigure[CGT]{
    \begin{minipage}[b]{0.12\textwidth}
    \includegraphics[width=1.0\linewidth]{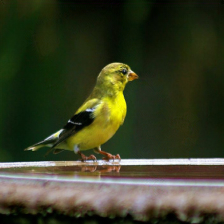}
    \includegraphics[width=1.0\linewidth]{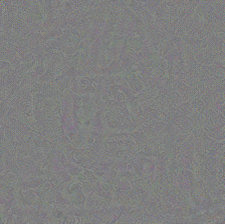}
    \includegraphics[width=1.0\linewidth]{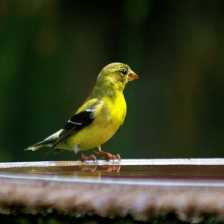}
    \includegraphics[width=1.0\linewidth]{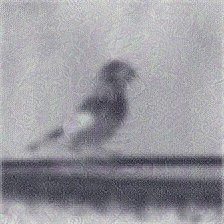}
    \includegraphics[width=1.0\linewidth]{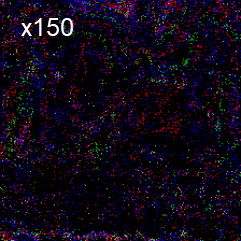}
    \includegraphics[width=1.0\linewidth]{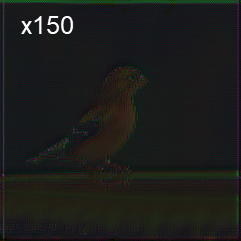}
    \end{minipage}
    }
    \caption{Visualization of $\bm{x}_{cln}$, $\bm{x}_{tgt}$, $\bm{x}_{adv}$, $\bm{x}_{r}$, $| \bm{x}_{adv}-\bm{x}_{cln} |$, and the estimated dropped information with different target images.}
    \label{fig:analysis}
\end{figure}
% -----------------------------------------------------------------------------

The Invertible Information Exchange Module (IIEM) is with the information preservation property, and performs feature-level information exchanging between the input clean image and the target image. Table~\ref{tab:effectiveness of IIEM} shows the performance of AdvINN with IIEM, without IIEM, and using a CNN to replace IIEM. When IIEM is not used in AdvINN, the adversarial examples are generated by directly combining of the benign image and the learned target image. From Table \ref{tab:effectiveness of IIEM}, we can observe that the results of w/o IIEM are with a significant deterioration compared to those of w/ IIEM. 
This indicates that IIEM is an indispensable component in AdvINN and can improve the image quality and accelerate convergence. In Table~\ref{tab:effectiveness of IIEM}, w/ CNN denotes that IIEM is replaced by a CNN~\cite{xiao2018generating}. We can see that the scores of all metrics further deteriorate except the FID score, moreover it takes much more iterations to converge. This result confirms that the information preservation property of IIEM is essential to the success of AdvINN. 

\subsection{Visualization and Analysis}

Fig. \ref{fig:analysis} visualizes input images, output images, adversarial perturbations and the estimated dropped information when using different target images. The $1$-st row shows the input clean images with the class label \textit{goldfinch}.
In the $2$-nd row, we show the target images with the highest confidence of the \textit{face powder}, the targeted UAP generated by \cite{NEURIPS2021_30d454f0} and the classifier guided target image.
From the output adversarial examples $\bm{x}_{adv}$ in the third row, we cannot see noticeable visual differences compared to the input clean image in all cases. This further verifies the effectiveness of the proposed method.
In order to further interpret the results of AdvINN, we visualizes the residual images $\bm{x}_{r}$ ($\bm{x}_{r}$ is normalized for clearer perception), the absolute difference between $\bm{x}_{cln}$ and $\bm{x}_{tgt}$, and the estimated dropped information\footnote{The dropped information is estimated by replacing the target image with a constant image (with no information) while keeping parameters of AdvINN fixed, therefore the generated residual image only contains the dropped information from the clean image.} in row $4$ to row $6$, respectively. We can see that the adversarial example generated by AdvINN-HCT contains the boundary information of the target image and discards the discriminant high-frequency features of the goldfinch; the adversarial example crafted by AdvINN-UAP includes some universal adversarial perturbation patterns and drops certain key features corresponding to head, chest and tail of the goldfinch; the adversarial example generated by AdvINN-CGT only adds minor modification to the clean images which is enlarged by 150 times for better visualization, and drops slight information corresponding to the shape of the goldfinch.

In summary, we can observe that AdvINN drops discriminant information (high-frequency details or shape information) of clean images and adds class-specific information from the target images simultaneously. 

\section{Ablation Study}
\subsection{Adversarial Budget $\epsilon$}
The adversarial budget controls the maximum amplitude of the  perturbation allowed on the generated adversarial examples. The performance of certain adversarial attack methods would be limited if a smaller adversarial budget is required.
Table \ref{tab:three constraints} shows the performance of AdvINN with three different adversarial budgets, \textit{i.e.}, 4/255, 8/255, and 16/255. 
We can see that there is no significant difference on the performance of AdvINN under the different constraints. This indicates that the quality of the adversarial examples generated by AdvINN does not limited by the maximum perturbation constraint, and AdvINN method maintains a stable convergence speed and achieves high attacking success rates even under a stricter perturbation budget.
For fair comparisons, the adversarial budget $\epsilon$ is still set to 8/255 by $l_{\infty}$-norm which is consistent with other comparison methods.

\begin{table}[t]
  \centering
  \caption{Ablation study: the performance of AdvINN under different adversarial budget constraints.}
    \begin{tabular}{c|cccccc}
    \toprule
          $\epsilon$ & $l_{\infty}\downarrow$   %& PSNR$\uparrow$  
          % & $l_{\infty}\downarrow$    
          & LPIPS$\downarrow$  & FID$\downarrow$  & Iter$\downarrow$& ASR(\%)$\uparrow$\\
    \midrule
    \midrule
    4/255 &0.0172  %& 46.49  
    % & 4.57  & 0.02 
    & 0.0118 & 1.575 &341&100.0\\
    8/255 &0.0281 %& 46.38  
    % & 3.70  & 0.03 
    & 0.0118 & 1.594 &321&100.0\\
    16/255 &0.0332  %& 46.37  
    % & 3.73  & 0.04 
    & 0.0119 & 1.568 &325&100.0\\
    \bottomrule
    \end{tabular}%
  \label{tab:three constraints}%
\end{table}%

% \subsection{Number of DWT/IDWT Layers} 
% Discrete Wavelet Transform (DWT) decomposes the input image into low-frequency and high-frequency components and at the same time decreases the spatial dimension of the feature and increases the number of feature channels. It can be applied multiple times for further decomposition. As shown in Table \ref{tab:effectiveness of DWT}, we can see that if DWT is applied two times, it can moderately improve the optimization speed from 321 to 271, while image quality would slightly deteriorate. 
% Therefore, for ImageNet-1K dataset, we apply 1 pair of DWT/IDWT for decomposition and reconstruction. 
% % The same convergence results can be observed on CIFAR-10 and CIFAR-100.

% \begin{table}[htbp]
%   \centering
%   \caption{Ablation study: the effect of different DWT/IDWT layers evaluated on ImageNet-1K dataset.}
%     \begin{tabular}{c|cccccc}
%     \toprule
%           DWT & $l_{2}\downarrow$ & SSIM$\uparrow$  %& PSNR$\uparrow$  
%               & LPIPS$\downarrow$ & FID$\downarrow$   & Iter$\downarrow$ \\
%     \midrule
%     \midrule
%     1 & 2.66 & 0.996 & 0.0118 & 1.594 & 321\\
%     2 & 2.71 & 0.995 & 0.0124 & 1.866 & 271\\% & 46.00 
%     \bottomrule
%     \end{tabular}%
%   \label{tab:effectiveness of DWT}%
% \end{table}%

\subsection{Results on other classifiers} 

\begin{table}[t]
  \centering
  \caption{The performance of AdvINN on different classifiers. The adversarial weights $\mathcal{\lambda}_{adv}$ are set to 10 and 3 on Inception\_v3 and  Densenet121, respectively. }
    \begin{tabular}{l|ccccc}
    \toprule
    Classifier & $l_{2}\downarrow$  %& PSNR $\uparrow$  & SSIM $\uparrow$ 
        & LPIPS$\downarrow$ & FID$\downarrow$ &ASR(\%)$\uparrow$\\
    \midrule
    \midrule
    Inception\_v3  & 4.57  & 0.0155 & 2.600 &100.0\\%& 0.992
    Densenet121    & 2.51  & 0.0114 & 1.604  &100.0\\% & 47.80 & 0.996
    \bottomrule
    \end{tabular}%
  \label{tab:classifier}%
\end{table}%

We have also tested AdvINN on the other two classifiers: Densenet121~\cite{huang2017densely} and Inception\_v3~\cite{szegedy2016rethinking}. Densenet121 fully utilizes the features by dense connection, which largely reduces the number of parameters. But the reduction in parameters leads to weaker robustness against adversarial attacks. Inception\_v3 utilizes convolution kernels of different sizes and is more complex in structure, but more robust against adversarial attacks. 

We adjust the adversarial weights $\mathcal{\lambda}_{adv}$ on different classifiers for better performance. Specifically, $\mathcal{\lambda}_{adv}$ is set to 10 on Inception\_v3 and 3 on Densenet121, respectively. Table \ref{tab:classifier} shows the experimental results and with both conditions, AdvINN achieves 100\% success attacking rate. We observe that AdvINN can be simply applied to other classifiers. Even with a more robust classifier, AdvINN succeeds in generating more imperceptible adversarial examples.
\section{Conclusion}
\label{sec:conclusion}
In this paper, we propose a novel adversarial attack framework, termed as AdvINN, to generate adversarial examples based on Invertible Neural Networks (INNs). 
By utilizing the information preservation property of INNs, the proposed Invertible Information Exchange Module, driven by the loss functions, performs information exchanging at the feature level and achieves simultaneously dropping discriminant information of clean images and adding class-specific features of the target images to craft adversaries. Moreover, three target image selection and learning approaches have been carefully investigated and analyzed.
Extensive experimental results have shown that the proposed AdvINN method can generate visually less perceptible and more robust adversarial examples compared to the state-of-the-art methods. 

\section{Acknowledgments}
This work is supported by the National Natural Science Foundation of China under Project 62201600 and U1811462, and NUDT Research Project ZK22-56.

% Use \bibliography{yourbibfile} instead or the References section will not appear in your paper
% \input{aaai23.bbl}
\bibliography{aaai23}

\begin{thebibliography}{58}
\providecommand{\natexlab}[1]{#1}

\bibitem[{Akhtar and Mian(2018)}]{Akhtar2018ThreatOA}
Akhtar, N.; and Mian, A.~S. 2018.
\newblock Threat of Adversarial Attacks on Deep Learning in Computer Vision: A
  Survey.
\newblock \emph{IEEE Access}, 6: 14410--14430.

\bibitem[{Ardizzone et~al.(2019)Ardizzone, L{\"u}th, Kruse, Rother, and
  K{\"o}the}]{ardizzone2019guided}
Ardizzone, L.; L{\"u}th, C.; Kruse, J.; Rother, C.; and K{\"o}the, U. 2019.
\newblock Guided image generation with conditional invertible neural networks.
\newblock \emph{arXiv preprint arXiv:1907.02392}.

\bibitem[{Benz et~al.(2020)Benz, Zhang, Imtiaz, and Kweon}]{benz2020double}
Benz, P.; Zhang, C.; Imtiaz, T.; and Kweon, I.~S. 2020.
\newblock Double targeted universal adversarial perturbations.
\newblock In \emph{Proceedings of the Asian Conference on Computer Vision}.

\bibitem[{Carlini and Wagner(2017)}]{carlini2017towards}
Carlini, N.; and Wagner, D. 2017.
\newblock Towards evaluating the robustness of neural networks.
\newblock In \emph{2017 ieee symposium on security and privacy (sp)}, 39--57.
  IEEE.

\bibitem[{Cheng, Xie, and Chen(2021)}]{cheng2021iicnet}
Cheng, K.~L.; Xie, Y.; and Chen, Q. 2021.
\newblock Iicnet: A generic framework for reversible image conversion.
\newblock In \emph{Proceedings of the IEEE/CVF International Conference on
  Computer Vision}, 1991--2000.

\bibitem[{Croce and Hein(2019)}]{croce2019sparse}
Croce, F.; and Hein, M. 2019.
\newblock Sparse and imperceivable adversarial attacks.
\newblock In \emph{Proceedings of the IEEE/CVF International Conference on
  Computer Vision}, 4724--4732.

\bibitem[{Das et~al.(2018)Das, Shanbhogue, Chen, Hohman, Li, Chen, Kounavis,
  and Chau}]{2018Shield}
Das, N.; Shanbhogue, M.; Chen, S.~T.; Hohman, F.; Li, S.; Chen, L.; Kounavis,
  M.~E.; and Chau, D.~H. 2018.
\newblock Shield: Fast, Practical Defense and Vaccination for Deep Learning
  using JPEG Compression.
\newblock In \emph{the 24th ACM SIGKDD International Conference}.

\bibitem[{Dinh, Krueger, and Bengio(2014)}]{dinh2014nice}
Dinh, L.; Krueger, D.; and Bengio, Y. 2014.
\newblock Nice: Non-linear independent components estimation.
\newblock \emph{arXiv preprint arXiv:1410.8516}.

\bibitem[{Dinh, Sohl-Dickstein, and Bengio(2016)}]{dinh2016density}
Dinh, L.; Sohl-Dickstein, J.; and Bengio, S. 2016.
\newblock Density estimation using real {NVP}.
\newblock \emph{arXiv preprint arXiv:1605.08803}.

\bibitem[{Dong et~al.(2018)Dong, Liao, Pang, Su, Zhu, Hu, and
  Li}]{dong2018boosting}
Dong, Y.; Liao, F.; Pang, T.; Su, H.; Zhu, J.; Hu, X.; and Li, J. 2018.
\newblock Boosting adversarial attacks with momentum.
\newblock In \emph{Proceedings of the IEEE conference on computer vision and
  pattern recognition}, 9185--9193.

\bibitem[{Duan et~al.(2021)Duan, Chen, Niu, Yang, Qin, and
  He}]{duan2021advdrop}
Duan, R.; Chen, Y.; Niu, D.; Yang, Y.; Qin, A.~K.; and He, Y. 2021.
\newblock AdvDrop: Adversarial Attack to DNNs by Dropping Information.
\newblock In \emph{Proceedings of the IEEE/CVF International Conference on
  Computer Vision}, 7506--7515.

\bibitem[{Goodfellow, Shlens, and Szegedy(2014)}]{goodfellow2014explaining}
Goodfellow, I.~J.; Shlens, J.; and Szegedy, C. 2014.
\newblock Explaining and harnessing adversarial examples.
\newblock \emph{arXiv preprint arXiv:1412.6572}.

\bibitem[{Guan et~al.(2022)Guan, Jing, Deng, Xu, Jiang, Zhang, and
  Li}]{guan2022deepmih}
Guan, Z.; Jing, J.; Deng, X.; Xu, M.; Jiang, L.; Zhang, Z.; and Li, Y. 2022.
\newblock DeepMIH: Deep Invertible Network for Multiple Image Hiding.
\newblock \emph{IEEE Transactions on Pattern Analysis and Machine
  Intelligence}.

\bibitem[{Guo et~al.(2017)Guo, Rana, Cisse, and Laurens}]{2017Countering}
Guo, C.; Rana, M.; Cisse, M.; and Laurens, V. 2017.
\newblock Countering Adversarial Images using Input Transformations.

\bibitem[{Hendrycks et~al.(2021)Hendrycks, Zhao, Basart, Steinhardt, and
  Song}]{Hendrycks2021NaturalAE}
Hendrycks, D.; Zhao, K.; Basart, S.; Steinhardt, J.; and Song, D.~X. 2021.
\newblock Natural Adversarial Examples.
\newblock \emph{2021 IEEE/CVF Conference on Computer Vision and Pattern
  Recognition (CVPR)}, 15257--15266.

\bibitem[{Heusel et~al.(2017)Heusel, Ramsauer, Unterthiner, Nessler, and
  Hochreiter}]{heusel2017gans}
Heusel, M.; Ramsauer, H.; Unterthiner, T.; Nessler, B.; and Hochreiter, S.
  2017.
\newblock Gans trained by a two time-scale update rule converge to a local nash
  equilibrium.
\newblock \emph{Advances in neural information processing systems}, 30.

\bibitem[{Huang et~al.(2017)Huang, Liu, Van Der~Maaten, and
  Weinberger}]{huang2017densely}
Huang, G.; Liu, Z.; Van Der~Maaten, L.; and Weinberger, K.~Q. 2017.
\newblock Densely connected convolutional networks.
\newblock In \emph{Proceedings of the IEEE conference on computer vision and
  pattern recognition}, 4700--4708.

\bibitem[{Huang and Dragotti(2021)}]{huang2021linn}
Huang, J.-J.; and Dragotti, P.~L. 2021.
\newblock LINN: Lifting inspired invertible neural network for image denoising.
\newblock In \emph{2021 29th European Signal Processing Conference (EUSIPCO)},
  636--640. IEEE.

\bibitem[{Huang and Dragotti(2022)}]{huang2021winnet}
Huang, J.-J.; and Dragotti, P.~L. 2022.
\newblock WINNet: Wavelet-Inspired Invertible Network for Image Denoising.
\newblock \emph{IEEE Transactions on Image Processing}, 31: 4377--4392.

\bibitem[{Huang et~al.(2022)Huang, Liu, Yang, Fu, Zhao, and
  Dragotti}]{huang2022durrnet}
Huang, J.-J.; Liu, T.; Yang, Z.; Fu, S.; Zhao, W.; and Dragotti, P.~L. 2022.
\newblock DURRNet: Deep Unfolded Single Image Reflection Removal Network.
\newblock \emph{arXiv preprint arXiv:2203.06306}.

\bibitem[{Huang et~al.(2021)Huang, Chen, Lu, Wang, Peng, and
  Huang}]{huang2021video}
Huang, Y.-C.; Chen, Y.-H.; Lu, C.-Y.; Wang, H.-P.; Peng, W.-H.; and Huang,
  C.-C. 2021.
\newblock Video Rescaling Networks with Joint Optimization Strategies for
  Downscaling and Upscaling.
\newblock In \emph{Proceedings of the IEEE/CVF Conference on Computer Vision
  and Pattern Recognition}, 3527--3536.

\bibitem[{Jacobsen, Smeulders, and Oyallon(2018)}]{jacobsen2018revnet}
Jacobsen, J.-H.; Smeulders, A.; and Oyallon, E. 2018.
\newblock i-RevNet: Deep Invertible Networks.
\newblock In \emph{ICLR 2018-International Conference on Learning
  Representations}.

\bibitem[{Jia et~al.(2022)Jia, Ma, Yao, Yin, Ding, and Yang}]{jia2022exploring}
Jia, S.; Ma, C.; Yao, T.; Yin, B.; Ding, S.; and Yang, X. 2022.
\newblock Exploring Frequency Adversarial Attacks for Face Forgery Detection.
\newblock In \emph{Proceedings of the IEEE/CVF Conference on Computer Vision
  and Pattern Recognition}, 4103--4112.

\bibitem[{Jing et~al.(2021)Jing, Deng, Xu, Wang, and Guan}]{Jing2021HiNetDI}
Jing, J.; Deng, X.; Xu, M.; Wang, J.; and Guan, Z. 2021.
\newblock HiNet: Deep Image Hiding by Invertible Network.
\newblock \emph{2021 IEEE/CVF International Conference on Computer Vision
  (ICCV)}, 4713--4722.

\bibitem[{Khrulkov and Oseledets(2018)}]{Khrulkov2018ArtOS}
Khrulkov, V.; and Oseledets, I. 2018.
\newblock Art of Singular Vectors and Universal Adversarial Perturbations.
\newblock \emph{2018 IEEE/CVF Conference on Computer Vision and Pattern
  Recognition}, 8562--8570.

\bibitem[{Kingma and Ba(2014)}]{kingma2014adam}
Kingma, D.~P.; and Ba, J. 2014.
\newblock Adam: A method for stochastic optimization.
\newblock \emph{arXiv preprint arXiv:1412.6980}.

\bibitem[{Kingma and Dhariwal(2018)}]{kingma2018glow}
Kingma, D.~P.; and Dhariwal, P. 2018.
\newblock Glow: Generative flow with invertible 1x1 convolutions.

\bibitem[{Kurakin, Goodfellow, and Bengio(2016)}]{kurakin2016adversarial}
Kurakin, A.; Goodfellow, I.; and Bengio, S. 2016.
\newblock Adversarial machine learning at scale.
\newblock \emph{arXiv preprint arXiv:1611.01236}.

\bibitem[{Lin et~al.(2019)Lin, Song, He, Wang, and Hopcroft}]{lin2019nesterov}
Lin, J.; Song, C.; He, K.; Wang, L.; and Hopcroft, J.~E. 2019.
\newblock Nesterov Accelerated Gradient and Scale Invariance for Adversarial
  Attacks.
\newblock In \emph{International Conference on Learning Representations}.

\bibitem[{Liu et~al.(2021)Liu, Qin, Anwar, Ji, Kim, Caldwell, and
  Gedeon}]{Liu_2021_CVPR}
Liu, Y.; Qin, Z.; Anwar, S.; Ji, P.; Kim, D.; Caldwell, S.; and Gedeon, T.
  2021.
\newblock Invertible Denoising Network: A Light Solution for Real Noise
  Removal.
\newblock In \emph{Proceedings of the IEEE/CVF Conference on Computer Vision
  and Pattern Recognition (CVPR)}, 13365--13374.

\bibitem[{Lu et~al.(2021)Lu, Wang, Zhong, and Rosin}]{lu2021large}
Lu, S.-P.; Wang, R.; Zhong, T.; and Rosin, P.~L. 2021.
\newblock Large-capacity image steganography based on invertible neural
  networks.
\newblock In \emph{Proceedings of the IEEE/CVF Conference on Computer Vision
  and Pattern Recognition}, 10816--10825.

\bibitem[{Luo et~al.(2018)Luo, Liu, Wei, and Xu}]{luo2018towards}
Luo, B.; Liu, Y.; Wei, L.; and Xu, Q. 2018.
\newblock Towards imperceptible and robust adversarial example attacks against
  neural networks.
\newblock In \emph{Proceedings of the AAAI Conference on Artificial
  Intelligence}, volume~32.

\bibitem[{Luo et~al.(2022)Luo, Lin, Xie, Wu, Xie, and Shen}]{luo2022frequency}
Luo, C.; Lin, Q.; Xie, W.; Wu, B.; Xie, J.; and Shen, L. 2022.
\newblock Frequency-driven Imperceptible Adversarial Attack on Semantic
  Similarity.
\newblock In \emph{Proceedings of the IEEE/CVF Conference on Computer Vision
  and Pattern Recognition}, 15315--15324.

\bibitem[{Madry et~al.(2018)Madry, Makelov, Schmidt, Tsipras, and
  Vladu}]{madry2018towards}
Madry, A.; Makelov, A.; Schmidt, L.; Tsipras, D.; and Vladu, A. 2018.
\newblock Towards Deep Learning Models Resistant to Adversarial Attacks.
\newblock In \emph{International Conference on Learning Representations}.

\bibitem[{Mallat(1989)}]{mallat1989theory}
Mallat, S.~G. 1989.
\newblock A theory for multiresolution signal decomposition: the wavelet
  representation.
\newblock \emph{IEEE transactions on pattern analysis and machine
  intelligence}, 11(7): 674--693.

\bibitem[{Mohaghegh~Dolatabadi, Erfani, and
  Leckie(2020)}]{mohaghegh2020advflow}
Mohaghegh~Dolatabadi, H.; Erfani, S.; and Leckie, C. 2020.
\newblock Advflow: Inconspicuous black-box adversarial attacks using
  normalizing flows.
\newblock \emph{Advances in Neural Information Processing Systems}, 33:
  15871--15884.

\bibitem[{Moosavi-Dezfooli et~al.(2017)Moosavi-Dezfooli, Fawzi, Fawzi, and
  Frossard}]{UAPs2017}
Moosavi-Dezfooli, S.-M.; Fawzi, A.; Fawzi, O.; and Frossard, P. 2017.
\newblock Universal Adversarial Perturbations.
\newblock In \emph{2017 IEEE Conference on Computer Vision and Pattern
  Recognition (CVPR)}.

\bibitem[{Moosavi-Dezfooli, Fawzi, and Frossard(2016)}]{moosavi2016deepfool}
Moosavi-Dezfooli, S.-M.; Fawzi, A.; and Frossard, P. 2016.
\newblock Deepfool: a simple and accurate method to fool deep neural networks.
\newblock In \emph{Proceedings of the IEEE conference on computer vision and
  pattern recognition}, 2574--2582.

\bibitem[{Naseer et~al.(2020)Naseer, Khan, Hayat, Khan, and
  Porikli}]{naseer2020self}
Naseer, M.; Khan, S.; Hayat, M.; Khan, F.~S.; and Porikli, F. 2020.
\newblock A self-supervised approach for adversarial robustness.
\newblock In \emph{Proceedings of the IEEE/CVF Conference on Computer Vision
  and Pattern Recognition}, 262--271.

\bibitem[{Poursaeed et~al.(2018)Poursaeed, Katsman, Gao, and
  Belongie}]{poursaeed2018generative}
Poursaeed, O.; Katsman, I.; Gao, B.; and Belongie, S. 2018.
\newblock Generative adversarial perturbations.
\newblock In \emph{Proceedings of the IEEE Conference on Computer Vision and
  Pattern Recognition}, 4422--4431.

\bibitem[{Russakovsky et~al.(2015)Russakovsky, Deng, Su, Krause, Satheesh, Ma,
  Huang, Karpathy, Khosla, Bernstein et~al.}]{russakovsky2015imagenet}
Russakovsky, O.; Deng, J.; Su, H.; Krause, J.; Satheesh, S.; Ma, S.; Huang, Z.;
  Karpathy, A.; Khosla, A.; Bernstein, M.; et~al. 2015.
\newblock Imagenet large scale visual recognition challenge.
\newblock \emph{International journal of computer vision}, 115(3): 211--252.

\bibitem[{Szegedy et~al.(2016)Szegedy, Vanhoucke, Ioffe, Shlens, and
  Wojna}]{szegedy2016rethinking}
Szegedy, C.; Vanhoucke, V.; Ioffe, S.; Shlens, J.; and Wojna, Z. 2016.
\newblock Rethinking the inception architecture for computer vision.
\newblock In \emph{Proceedings of the IEEE conference on computer vision and
  pattern recognition}, 2818--2826.

\bibitem[{Szegedy et~al.(2014)Szegedy, Zaremba, Sutskever, Bruna, Erhan,
  Goodfellow, and Fergus}]{DBLP:journals/corr/SzegedyZSBEGF13}
Szegedy, C.; Zaremba, W.; Sutskever, I.; Bruna, J.; Erhan, D.; Goodfellow,
  I.~J.; and Fergus, R. 2014.
\newblock Intriguing properties of neural networks.
\newblock In Bengio, Y.; and LeCun, Y., eds., \emph{2nd International
  Conference on Learning Representations, {ICLR} 2014, Banff, AB, Canada, April
  14-16, 2014, Conference Track Proceedings}.

\bibitem[{Tian et~al.(2022)Tian, Pan, Yang, Zhang, He, and
  Jin}]{tian2022imperceptible}
Tian, Y.; Pan, J.; Yang, S.; Zhang, X.; He, S.; and Jin, Y. 2022.
\newblock Imperceptible and Sparse Adversarial Attacks via a Dual-Population
  Based Constrained Evolutionary Algorithm.
\newblock \emph{IEEE Transactions on Artificial Intelligence}.

\bibitem[{Wang et~al.(2021)Wang, He, Wang, and He}]{wang2021admix}
Wang, X.; He, X.; Wang, J.; and He, K. 2021.
\newblock Admix: Enhancing the transferability of adversarial attacks.
\newblock In \emph{Proceedings of the IEEE/CVF International Conference on
  Computer Vision}, 16158--16167.

\bibitem[{Wang et~al.(2018)Wang, Yu, Wu, Gu, Liu, Dong, Qiao, and
  Change~Loy}]{wang2018esrgan}
Wang, X.; Yu, K.; Wu, S.; Gu, J.; Liu, Y.; Dong, C.; Qiao, Y.; and Change~Loy,
  C. 2018.
\newblock Esrgan: Enhanced super-resolution generative adversarial networks.
\newblock In \emph{Proceedings of the European conference on computer vision
  (ECCV) workshops}, 0--0.

\bibitem[{Wang et~al.(2004)Wang, Bovik, Sheikh, and Simoncelli}]{wang2004image}
Wang, Z.; Bovik, A.~C.; Sheikh, H.~R.; and Simoncelli, E.~P. 2004.
\newblock Image quality assessment: from error visibility to structural
  similarity.
\newblock \emph{IEEE transactions on image processing}, 13(4): 600--612.

\bibitem[{Xiao et~al.(2018)Xiao, Li, Zhu, He, Liu, and
  Song}]{xiao2018generating}
Xiao, C.; Li, B.; Zhu, J.-Y.; He, W.; Liu, M.; and Song, D. 2018.
\newblock Generating adversarial examples with adversarial networks.
\newblock \emph{arXiv preprint arXiv:1801.02610}.

\bibitem[{Xiao et~al.(2022)Xiao, Zheng, Liu, Lin, and Liu}]{xiao2022invertible}
Xiao, M.; Zheng, S.; Liu, C.; Lin, Z.; and Liu, T.-Y. 2022.
\newblock Invertible Rescaling Network and Its Extensions.
\newblock \emph{International Journal of Computer Vision}, 1--26.

\bibitem[{Xiao et~al.(2020)Xiao, Zheng, Liu, Wang, He, Ke, Bian, Lin, and
  Liu}]{xiao2020invertible}
Xiao, M.; Zheng, S.; Liu, C.; Wang, Y.; He, D.; Ke, G.; Bian, J.; Lin, Z.; and
  Liu, T.-Y. 2020.
\newblock Invertible image rescaling.
\newblock In \emph{Proceedings of European Conference on Computer Vision
  (ECCV)}, 126--144.

\bibitem[{Xu et~al.(2021)Xu, Wang, Wei, and Lu}]{xu2021compact}
Xu, H.-B.; Wang, R.; Wei, J.; and Lu, S.-P. 2021.
\newblock A Compact Neural Network-based Algorithm for Robust Image
  Watermarking.
\newblock \emph{arXiv e-prints}, arXiv--2112.

\bibitem[{Zhang et~al.(2020)Zhang, Benz, Imtiaz, and
  Kweon}]{zhang2020understanding}
Zhang, C.; Benz, P.; Imtiaz, T.; and Kweon, I.~S. 2020.
\newblock Understanding adversarial examples from the mutual influence of
  images and perturbations.
\newblock In \emph{Proceedings of the IEEE/CVF Conference on Computer Vision
  and Pattern Recognition}, 14521--14530.

\bibitem[{Zhang et~al.(2022)Zhang, Pan, Zhou, and Kuo}]{zhang2022enhancing}
Zhang, M.; Pan, Z.; Zhou, X.; and Kuo, C.-C.~J. 2022.
\newblock Enhancing Image Rescaling using Dual Latent Variables in Invertible
  Neural Network.
\newblock In \emph{Proceedings of the 30th ACM International Conference on
  Multimedia}, 5602--5610.

\bibitem[{Zhang et~al.(2018)Zhang, Isola, Efros, Shechtman, and
  Wang}]{zhang2018unreasonable}
Zhang, R.; Isola, P.; Efros, A.~A.; Shechtman, E.; and Wang, O. 2018.
\newblock The unreasonable effectiveness of deep features as a perceptual
  metric.
\newblock In \emph{Proceedings of the IEEE conference on computer vision and
  pattern recognition}, 586--595.

\bibitem[{Zhao et~al.(2021)Zhao, Liu, Xiao, Lun, and Lam}]{zhao2021invertible}
Zhao, R.; Liu, T.; Xiao, J.; Lun, D.~P.; and Lam, K.-M. 2021.
\newblock Invertible image decolorization.
\newblock \emph{IEEE Transactions on Image Processing}, 30: 6081--6095.

\bibitem[{Zhao, Liu, and Larson(2020)}]{zhao2020towards}
Zhao, Z.; Liu, Z.; and Larson, M. 2020.
\newblock Towards large yet imperceptible adversarial image perturbations with
  perceptual color distance.
\newblock In \emph{Proceedings of the IEEE/CVF Conference on Computer Vision
  and Pattern Recognition}, 1039--1048.

\bibitem[{Zhao, Liu, and Larson(2021)}]{NEURIPS2021_30d454f0}
Zhao, Z.; Liu, Z.; and Larson, M. 2021.
\newblock On Success and Simplicity: A Second Look at Transferable Targeted
  Attacks.
\newblock In Ranzato, M.; Beygelzimer, A.; Dauphin, Y.; Liang, P.; and Vaughan,
  J.~W., eds., \emph{Advances in Neural Information Processing Systems},
  volume~34, 6115--6128. Curran Associates, Inc.

\bibitem[{Zhu et~al.(2019)Zhu, Li, Zhang, Li, Liu, and Xue}]{zhu2019residual}
Zhu, X.; Li, Z.; Zhang, X.-Y.; Li, C.; Liu, Y.; and Xue, Z. 2019.
\newblock Residual invertible spatio-temporal network for video
  super-resolution.
\newblock In \emph{Proceedings of the AAAI conference on artificial
  intelligence}, volume~33, 5981--5988.

\end{thebibliography}

% AAAI is especially grateful to Peter Patel Schneider for his work in implementing the original aaai.sty file, liberally using the ideas of other style hackers, including Barbara Beeton. We also acknowledge with thanks the work of George Ferguson for his guide to using the style and BibTeX files --- which has been incorporated into this document --- and Hans Guesgen, who provided several timely modifications, as well as the many others who have, from time to time, sent in suggestions on improvements to the AAAI style. We are especially grateful to Francisco Cruz, Marc Pujol-Gonzalez, and Mico Loretan for the improvements to the Bib\TeX{} and \LaTeX{} files made in 2020.

% The preparation of the \LaTeX{} and Bib\TeX{} files that implement these instructions was supported by Schlumberger Palo Alto Research, AT\&T Bell Laboratories, Morgan Kaufmann Publishers, The Live Oak Press, LLC, and AAAI Press. Bibliography style changes were added by Sunil Issar. \verb+\+pubnote was added by J. Scott Penberthy. George Ferguson added support for printing the AAAI copyright slug. Additional changes to aaai23.sty and aaai23.bst have been made by Francisco Cruz, Marc Pujol-Gonzalez, and Mico Loretan.

% \bigskip
% \noindent Thank you for reading these instructions carefully. We look forward to receiving your electronic files!
\end{document}

% \maketitle
% \section{Supplementary}

\section{A. Algorithm of AdvINN}
The pseudo-code of the proposed AdvINN method is shown in \textbf{Algorithm 1} for reference:
% \begin{algorithm}[h]
%     \SetKwInOut{Input}{Input}\SetKwInOut{Output}{Output}
%     \caption{AdvINN-HCT and AdvINN-UAP}
%     \label{algorithm1}
%     \Input{ clean image $\bm{x}_{{cln}}$, target image  $\bm{x}_{{tgt}}$, adversarial budget $\epsilon$, confidence $\kappa$, learning rate $lr$; }%, the iteration number $t$}
%     % , the iteration numbers $t$
%     \Output{Adversarial image $\bm{x}_{{adv}}$;}
    
%     % \begin{algorithmic}[1] %[1] enables line numbers
    
%     Initialize the parameters of AdvINN: $\bm{\theta}$\; 
    
%     \While{$\bm{x}_{adv}$ \text{is not adversarial}}{
%         $(\bm{x}_{adv}, \bm{x}_{r}) \gets f_{\bm{\theta}}(\bm{x}_{cln}, \bm{x}_{tgt})$
%         % $\bm{x}_{{adv}} , r\gets \mathcal{I} (\bm{x}_{{cln}}, \bm{x}_{\text{target}})$\;
    
%         $\bm{x}_{{adv}}$ \gets $\text{min}(\bm{x}_{{cln}}+\epsilon, \text{max}(\bm{x}_{{adv}}, \bm{x}_{{cln}}-\epsilon))$\;
    
%         $p_{{tgt}} \gets$ $g(\bm{x}_{{adv}})$\;

%         \eIf{$p_{{tgt}} \leq \kappa$}{
%             Update loss function $\mathcal{L}_{total}$\;
        
%             Update $\bm{\theta} \gets \bm{\theta} + lr \cdot \text{Adam}(\mathcal{L}_{{total}})$\;
%         }
%         {
%             break\;
%         }
%     }
%     \textbf{return:} $\bm{x}_{{adv}}$.
% \end{algorithm}

\begin{algorithm}[h]
    \SetKwInOut{Input}{Input}\SetKwInOut{Output}{Output}
    \caption{AdvINN-CGT}
    \label{algorithm1}
    \Input{ clean image $\bm{x}_{{cln}}$, classifier guided image $\bm{x}_{{cgt}}$, adversarial budget $\epsilon$, confidence $\kappa$, learning rate $lr_{1}$, learning rate $lr_{2}$; }%, the iteration number $t$}
    % , the iteration numbers $t$
    \Output{Adversarial image $\bm{x}_{{adv}}$;}
    
    % \begin{algorithmic}[1] %[1] enables line numbers
    
    Initialize the parameters of AdvINN: $\bm{\theta}$\; 
    Initialize $\bm{x}_{{cgt}}$\ with all 0.5;
    % \footnote{$x_{cgt}$ takes 6 iterations to be able to fool the classifier on average. So we let $x_{cgt}$ synchronize with the update with $x_{adv}$};
    
    \While{$\bm{x}_{adv}$ \text{is not adversarial}}{
        % $(\bm{x}_{adv}, \bm{x}_{r}) \gets f_{\bm{\theta}}(\bm{x}_{cln}, \bm{x}_{tgt})$
        $(\bm{x}_{adv}, \bm{x}_{r}) \gets f_{\bm{\theta}}(\bm{x}_{cln}, \bm{x}_{cgt})$
        % $\bm{x}_{{adv}} , r\gets \mathcal{I} (\bm{x}_{{cln}}, \bm{x}_{\text{target}})$\;
    
        $\bm{x}_{{adv}}$ \gets $\text{min}(\bm{x}_{{cln}}+\epsilon, \text{max}(\bm{x}_{{adv}}, \bm{x}_{{cln}}-\epsilon))$\;
    
        $p_{{tgt}} \gets$ $g(\bm{x}_{{adv}})$\;
        % $p_{{cgt}} \gets$ $g(\bm{x}_{{cgt}})$\;

        \eIf{$p_{{tgt}} \leq \kappa$}{
            Update loss function $\mathcal{L}_{total}$\;
        
            Update $\bm{\theta} \gets \bm{\theta} + lr_{1} \cdot \text{Adam}(\mathcal{L}_{{total}})$\;
            
            Update loss function $\mathcal{L}_{cgt}$\;
            
            Update $\bm{x}_{{cgt}} \gets \bm{x}_{{cgt}} + lr_{2} \cdot \text{Adam}(\mathcal{L}_{{cgt}})$\;
        }
        {
            break\;
        }
    }
    \textbf{return:} $\bm{x}_{{adv}}$.
\end{algorithm}

\section{B. Ablation Study}

\subsection{B.1. Hyper-parameter $\mathcal{\lambda}_{adv}$}
The hyper-parameter $\mathcal{\lambda}_{adv}$ is used to set the trade-off between the image reconstruction objective and the adversarial objective.
Table \ref{tab:adv_wight} shows the performance of AdvINN with different $\mathcal{\lambda}_{adv}$. We can see that as $\mathcal{\lambda}_{adv}$ decreases, the quality of the generated adversarial image consistently improves, but the model requires more iterations to converge. 
If we further decrease $\mathcal{\lambda}_{adv}$, the attacking success rate would drop below 100\%. Specifically, when $\mathcal{\lambda}_{adv}$ equals to 2, the attacking success rate becomes 99.6\%. 
Therefore, $\mathcal{\lambda}_{adv}$ is set to 3 by considering the image quality, the convergence speed as well as the attacking success rate.
% Different $\mathcal{\lambda}_{adv}$ show diverse convergence speeds and image quality. If the $\mathcal{\lambda}_{adv}$ is set lower, the perceptual metrics will be improved. But it will correspondingly take more iterations to optimize. Considering the final quality and convergence speed, $\mathcal{\lambda}_{adv}$ is set to 3.

\begin{table}[t]
  \centering
  \caption{Ablation study: the performance of AdvINN evaluated on ImageNet-1K dataset with different adversarial weight $\mathcal{\lambda}_{adv}$. }
    \begin{tabular}{l|cccccc}
    \toprule
    $\mathcal{\lambda}_{adv}$ & $l_{2}\downarrow$ & SSIM $\uparrow$  %& PSNR $\uparrow$  
        & LPIPS$\downarrow$ & FID$\downarrow$ & Iter$\downarrow$ \\
    \midrule
    \midrule
    2     &2.41  & 0.996  & 0.0105 & 1.358 & 614 \\
    \textbf{3}    & 2.66  & 0.996  & 0.0118 & 1.594 & 321 \\% & 47.80
    5     & 3.08 & 0.995  & 0.0143 & 2.008 & 138 \\% & 47.22
    10    & 3.70 & 0.994  & 0.0189 & 2.900 & 65 \\% & 46.38
    15    & 4.19 & 0.994  & 0.0225 & 3.579 & 50 \\% & 45.82
    20    & 4.63 & 0.993   & 0.0258 & 4.222 & 39 \\% & 45.37
    25    & 5.05 & 0.992   & 0.0287 & 4.801 & 34 \\% & 44.99
    30    & 5.42 & 0.992  & 0.0312 & 5.311 & 31 \\% & 44.66
    40    & 6.14 & 0.991  & 0.0355 & 6.197 & 28 \\% & 44.10
    50    & 6.85 & 0.990  & 0.0393 & 6.940 & 26 \\% & 43.62
    60    & 7.54 & 0.989  & 0.0426 & 7.586 & 24 \\% & 43.21
    70    & 8.18 & 0.988  & 0.0455 & 8.186 & 23 \\% & 42.85
    80    & 8.87 & 0.987  & 0.0481 & 8.678 & 22 \\% & 42.50
    90    & 9.51 & 0.986  & 0.0505 & 9.173 & 21 \\% & 42.20
    100   & 10.15 & 0.986 & 0.0526 & 9.559 & 21 \\% & 41.93
    % 120   & 11.42 & 0.984 & 0.0564 & 10.314 & 20 \\% & 41.44
    % 150   & 13.17 & 0.981 & 0.0610 & 11.145 & 19 \\% & 40.83
    % 200   & 15.87 & 0.978 & 0.0668 & 12.213 & 19 \\% & 40.05
    % 400   & 23.37 & 0.969 & 0.0786 & 14.252 & 18 \\% & 38.35
    \bottomrule
    \end{tabular}%
  \label{tab:adv_wight}%
\end{table}%

\begin{table}[htbp]
  \centering
  \caption{Ablation study: the effect of different DWT/IDWT layers evaluated on ImageNet-1K dataset.}
    \begin{tabular}{c|cccccc}
    \toprule
          DWT & $l_{2}\downarrow$ & SSIM$\uparrow$  %& PSNR$\uparrow$  
              & LPIPS$\downarrow$ & FID$\downarrow$   & Iter$\downarrow$ \\
    \midrule
    \midrule
    1 & 2.66 & 0.996 & 0.0118 & 1.594 & 321\\
    2 & 2.71 & 0.995 & 0.0124 & 1.866 & 271\\% & 46.00 
    \bottomrule
    \end{tabular}%
  \label{tab:effectiveness of DWT}%
\end{table}%

\begin{table}[htbp]
  \centering
  \caption{Ablation study: the effect of the number of Affine Coupling Blocks on ImageNet-1K.}
  %The time represents generating adversaries successfully which reach 85\% confidence in 2,000 epochs on the whole dataset.}
    \begin{tabular}{c|ccccccc}
    \toprule
    Layer  & $l_{2}\downarrow$ & SSIM$\uparrow$ %& PSNR$\uparrow$  
        
    & LPIPS$\downarrow$ & FID$\downarrow$ & Iter$\downarrow$ \\%& Time$\downarrow$\\
    \midrule
    \midrule
    1     & \textbf{2.51} & \textbf{0.996}  & \textbf{0.0115} & 1.621 &418 \\%updated & \textbf{46.84} & 3906
    2     & 2.66 & \textbf{0.996}  & 0.0118 & \textbf{1.594}&\textbf{321} \\%updated & 46.38  & \textbf{3812}
    3     & 2.70 & \textbf{0.996} & 0.0121  & 1.653 & 323 \\%updated  & 46.27 & 4305
    4     & 2.73 & \textbf{0.996} & 0.0119  & 1.608 & 339\\%updated  & 46.19 & 4877
    \bottomrule
    \end{tabular}%
  \label{tab:layer of ACBs}%
\end{table}%

\subsection{B.2. Number of DWT/IDWT Layers} 
Discrete Wavelet Transform (DWT) decomposes the input image into low-frequency and high-frequency components and at the same time decreases the spatial dimension of the feature and increases the number of feature channels. It can be applied multiple times for further decomposition. As shown in Table \ref{tab:effectiveness of DWT}, we can see that if DWT is applied two times, it can moderately improve the optimization speed from 321 to 271, while image quality would slightly deteriorate. 
% But generating more noticeable perturbations than only once. 
Therefore, for ImageNet-1K dataset, we apply 1 pair of DWT/IDWT for decomposition and reconstruction. 

\subsection{B.3. Number of Affine Coupling Blocks}
The Affine Coupling Blocks perform invertible non-linear transformation on the features of the clean image and the target image. 
% therefore play a pivotal role in the proposed Invertible Information Exchange Module. 
Table \ref{tab:layer of ACBs} shows the effect of different number of Affine Coupling Blocks.
We can see that when the number is set to 1, the adversarial perturbations can achieve a slight better $l_2$ and LPIPS score, but the model will require 30\% more iterations than AdvINN with 2 Affine Coupling Blocks.
As the number of Affine Coupling Blocks increases (larger than 2), there would be a further increase on $l_2$ score and the model will require more iterations to converge. 
Therefore, by considering the image quality and the convergence speed, the number of Affine Coupling Blocks is set to 2 when evaluating the ImageNet-1K dataset.

\section{C. More Visual Results on ImageNet-1K} 

\begin{figure*}[t]
    \centering
    \includegraphics[width=\linewidth]{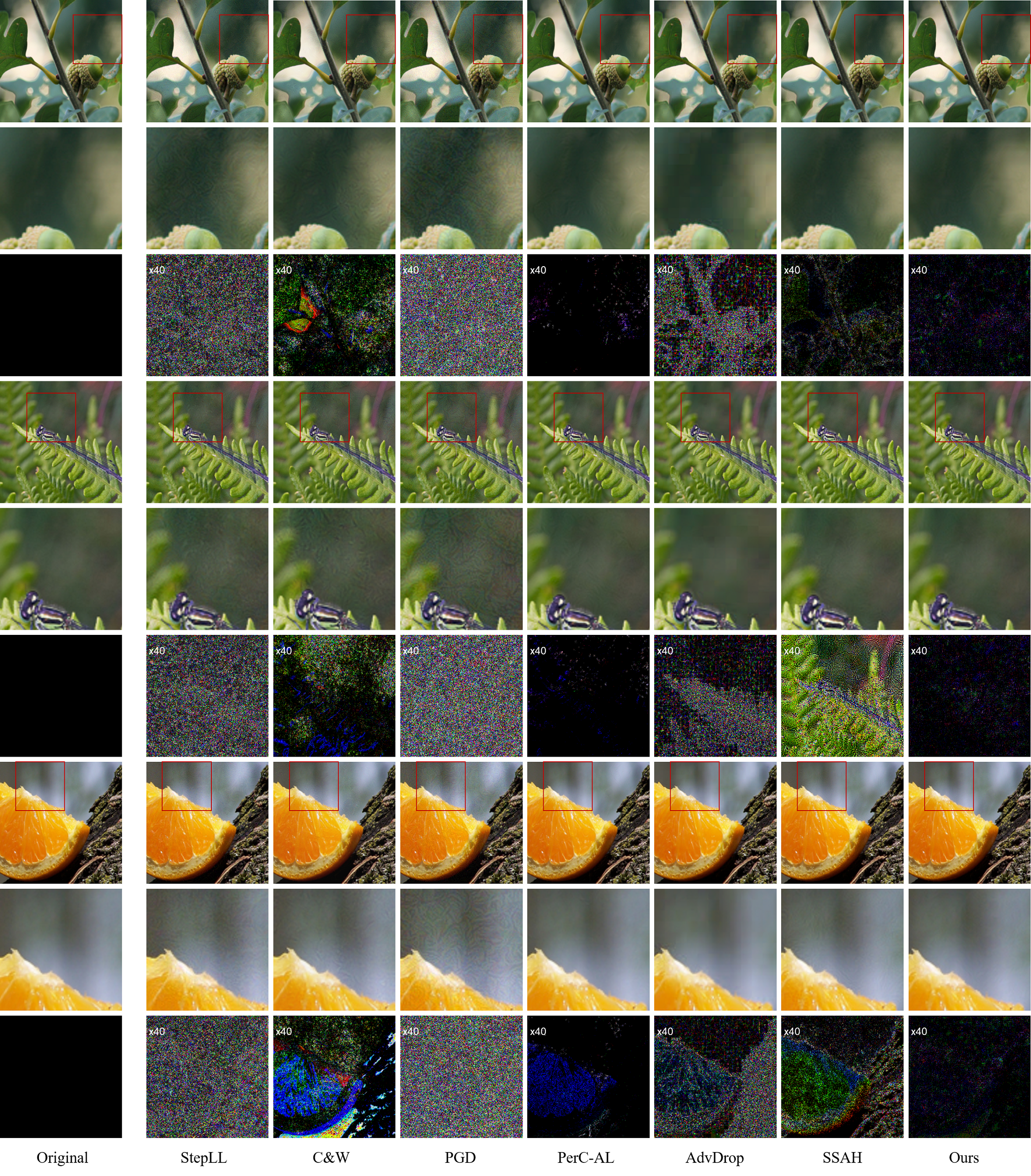}
    \caption{Adversarial examples and the corresponding adversarial perturbations generated by StepLL, C\&W, PGD, PerC-AL, AdvDrop, SSAH, and the proposed AdvINN method on ImageNet-1K. (Odd rows show the adversarial images, and even rows show the absolute value of adversarial perturbations magnified by 40 times.) }
    \label{fig:imagenet_1}
\end{figure*}

\begin{figure*}[t]
    \centering
    \includegraphics[width=\linewidth]{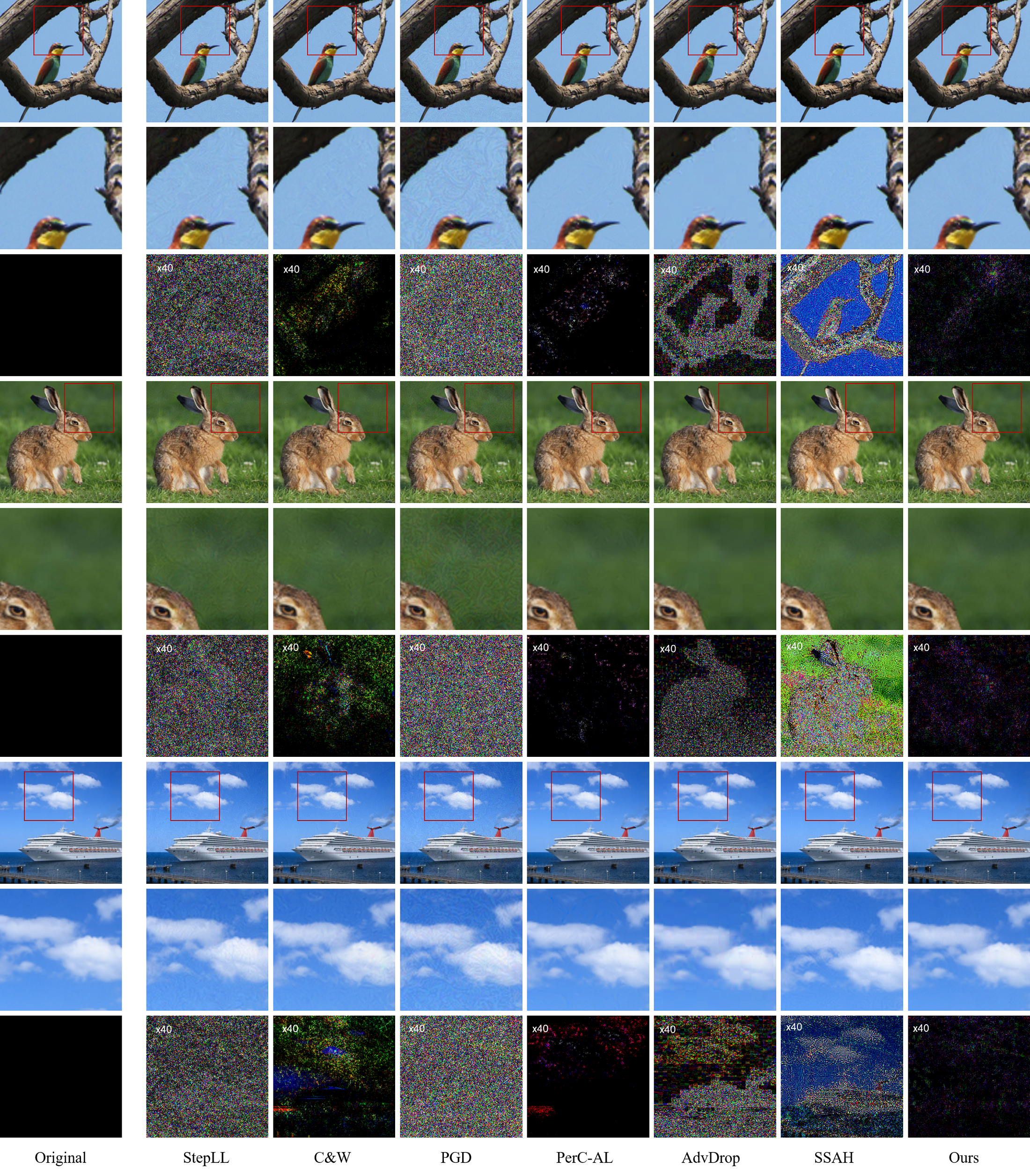}
    \caption{Adversarial examples and the corresponding adversarial perturbations generated by StepLL, C\&W, PGD, PerC-AL, AdvDrop, SSAH, and the proposed AdvINN method on ImageNet-1K. (Odd rows show the adversarial images, and even rows show the absolute value of adversarial perturbations magnified by 40 times.)}
    \label{fig:imagenet_2}
\end{figure*}

Fig. \ref{fig:imagenet_1} and Fig. \ref{fig:imagenet_2} show more exemplar adversarial examples and the corresponding adversarial perturbations generated by different methods. 

\begin{figure}[t]
    \centering
    \includegraphics[width=\linewidth]{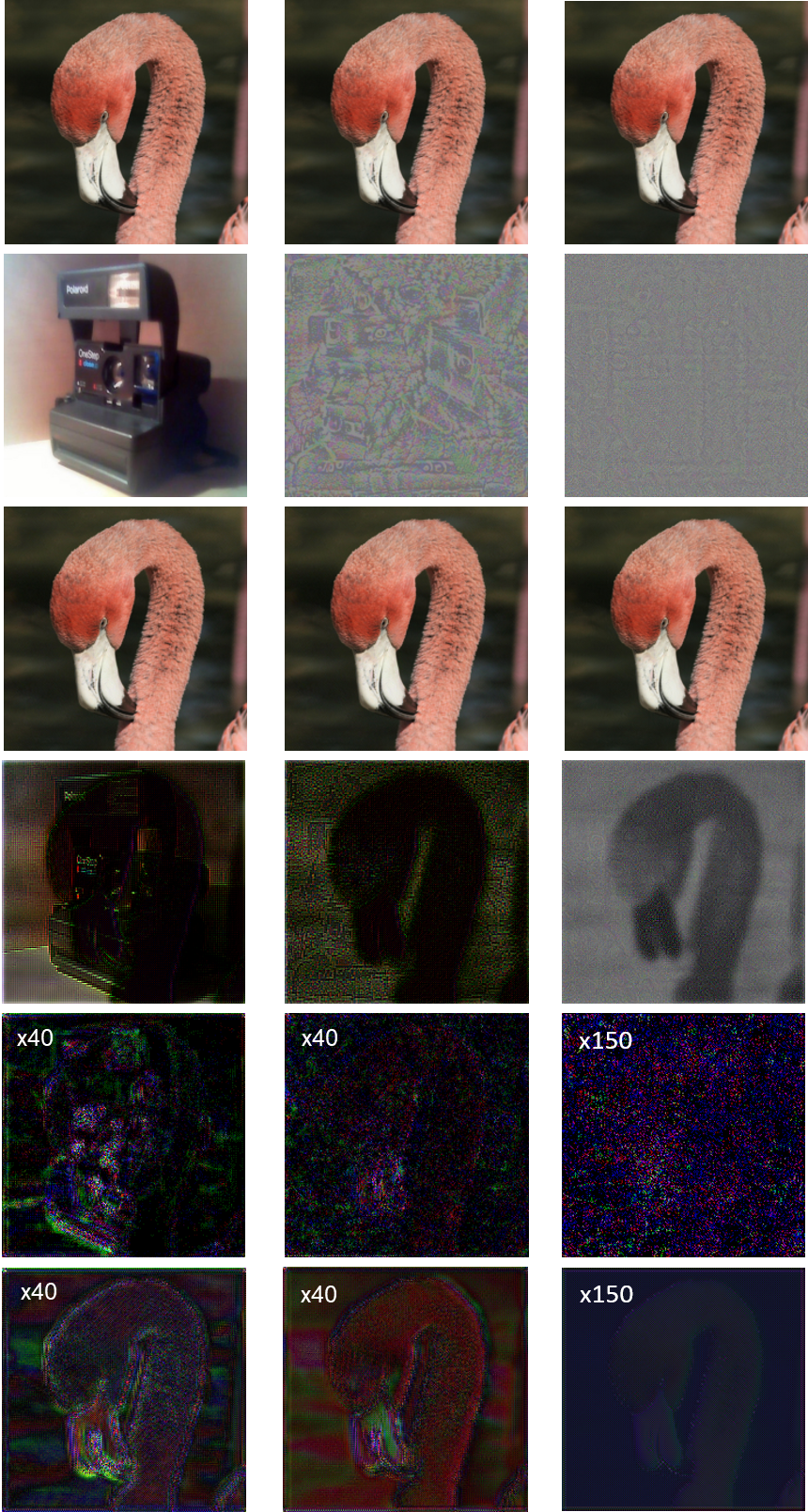}
    \caption{Visualization of $\bm{x}_{cln}$, $\bm{x}_{tgt}$, $\bm{x}_{adv}$, $\bm{x}_{r}$, $| \bm{x}_{adv}-\bm{x}_{cln} |$, and the estimated dropped information with different target images. The input clean image is of the \textit{flamingo} class and the target image is of the \textit{Polaroid camera} class.}
    \label{fig:analysis_exm2}
\end{figure}

\begin{figure}[t]
    \centering
    \includegraphics[width=0.985\linewidth]{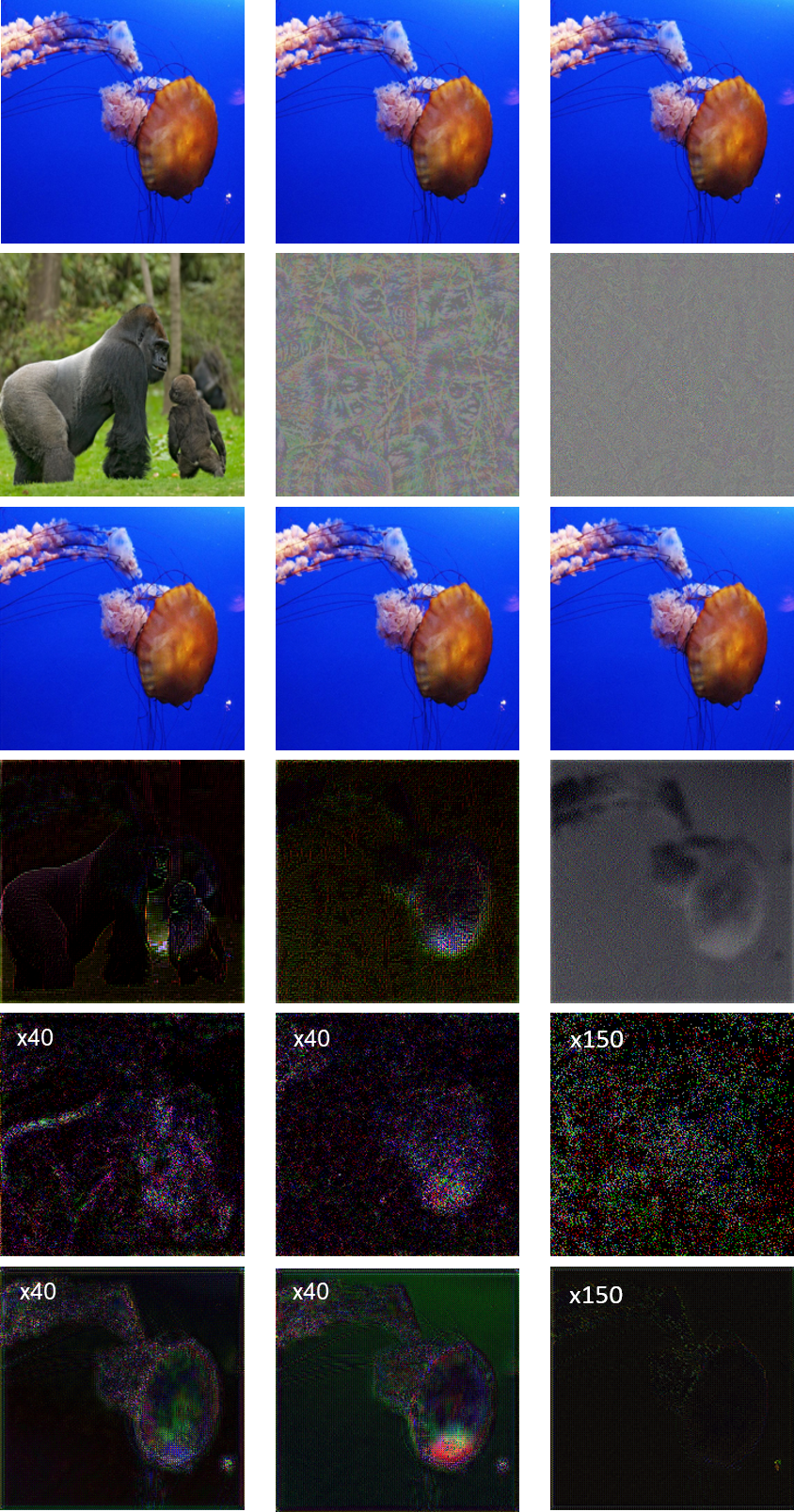}
    \caption{Visualization of $\bm{x}_{cln}$, $\bm{x}_{tgt}$, $\bm{x}_{adv}$, $\bm{x}_{r}$, $| \bm{x}_{adv}-\bm{x}_{cln} |$, and the estimated dropped information with different target images. The input clean image is of the \textit{jellyfish} class and the target image is of the \textit{gorilla} class.}
    \label{fig:analysis_exm3}
\end{figure}